\documentclass[conference]{IEEEtran}
\usepackage{times}

\usepackage[numbers]{natbib}
\usepackage{multicol}
\usepackage[bookmarks=true]{hyperref}

\usepackage[utf8]{inputenc} 
\usepackage[T1]{fontenc}    
\usepackage{hyperref}       
\usepackage{url}            
\usepackage{booktabs}       
\usepackage{amsfonts}       
\usepackage{nicefrac}       
\usepackage{microtype}      
\usepackage{graphicx}
\usepackage{rotating}
\usepackage{amsmath}
\let\labelindent\relax
\usepackage[inline]{enumitem}
\renewcommand{\baselinestretch}{0.98}
 
\newlist{paragraphs}{itemize*}{1}
\setlist[paragraphs]{
   label=(\textbf{\thesection}),
   itemjoin=\newline\hspace*{\parindent}
}
\makeatletter
\def\namedlabel#1#2#3{\begingroup
    #3%
    \def\@currentlabel{#2}%
    \phantomsection\label{#1}\endgroup
}
\makeatother
\usepackage{dsfont}
\usepackage{semicrunch}
\usepackage{wrapfig,lipsum,booktabs}
\usepackage{hyperref}

\usepackage{amsmath,amsfonts,bm}

\def\eqref#1{equation~\ref{#1}}

\def\1{\bm{1}}

\DeclareMathAlphabet{\mathsfit}{\encodingdefault}{\sfdefault}{m}{sl}
\SetMathAlphabet{\mathsfit}{bold}{\encodingdefault}{\sfdefault}{bx}{n}

\newcommand{\E}{\mathbb{E}}

\newcommand{\KL}{D_{\mathrm{KL}}}

\DeclareMathOperator*{\argmax}{arg\,max}
\DeclareMathOperator*{\argmin}{arg\,min}
\DeclareMathOperator*{\Ex}{\mathbb{E}}

\newcommand{\pinew}{{\pi_{k+1}}}
\newcommand{\piold}{{\pi_\beta}}

\newcommand{\buffer}{\beta}
\newcommand{\lagrangeawr}{\lambda}
\usepackage{xspace}
\usepackage[utf8]{inputenc}
\usepackage[english]{babel}
\usepackage{amsthm}
\usepackage{amssymb}
\usepackage{subcaption}
\usepackage{floatrow}
\usepackage{mathtools}
\usepackage{hyperref}
\usepackage{xcolor}

\newcommand\strike{\bgroup\markoverwith{\textcolor{red}{\rule[0.5ex]{2pt}{0.4pt}}}\ULon}
\definecolor{darkgreen}{RGB}{0,180,56}

\newcommand{\st}{\mathbf{s}}

\newcommand{\at}{\mathbf{a}}

\newcommand{\DTV}{D_\mathrm{TV}}

\newcommand{\METHOD}{Advantage Weighted Actor Critic\xspace}

\newtheorem*{rep@theorem}{\rep@title}
\newcommand{\newreptheorem}[2]{%
\newenvironment{rep#1}[1]{%
 \def\rep@title{#2 \ref{##1}}%
 \begin{rep@theorem}}%
 {\end{rep@theorem}}}
\makeatother
\newreptheorem{theorem}{Theorem}
\newreptheorem{lemma}{Lemma}
\newreptheorem{assumption}{Assumption}

\usepackage{graphicx}
\usepackage{wrapfig}
\usepackage{algorithm}
\usepackage{algcompatible}
\usepackage{multicol}
\usepackage{enumitem}

\usepackage{hyperref}
\usepackage{url}

\newcommand\extrafootertext[1]{%
    \bgroup
    \renewcommand\thefootnote{\fnsymbol{footnote}}%
    \renewcommand\thempfootnote{\fnsymbol{mpfootnote}}%
    \footnotetext[0]{#1}%
    \egroup
}

\newcommand\projectpage{\href{https://awacrl.github.io/}{\textit{awacrl.github.io}}}

\title{AWAC: Accelerating Online Reinforcement Learning with Offline Datasets}

\author{Ashvin Nair$^*$, Abhishek Gupta$^*$, Murtaza Dalal, Sergey Levine \\
Department of Electrical Engineering and Computer Science, UC Berkeley \\
}

\begin{document}

\maketitle 
\extrafootertext{$^*$Equal contribution. Correspondence to \texttt{nair@eecs.berkeley.edu}.}

\begin{abstract}
Reinforcement learning (RL) provides an appealing formalism for learning control policies from experience. However, the classic active formulation of RL necessitates a lengthy active exploration process for each behavior, making it difficult to apply in real-world settings such as robotic control. If we can instead allow RL algorithms to effectively use previously collected data to aid the online learning process, such applications could be made substantially more practical: the prior data would provide a starting point that mitigates challenges due to exploration and sample complexity, while the online training enables the agent to perfect the desired skill. Such prior data could either constitute expert demonstrations or, more generally, sub-optimal prior data that illustrates potentially useful transitions. While a number of prior methods have either used optimal demonstrations to bootstrap reinforcement learning, or have used sub-optimal data to train purely offline, it remains exceptionally difficult to train a policy with potentially sub-optimal offline data and actually continue to improve it further with online RL. In this paper we systematically analyze why this problem is so challenging, and propose an algorithm that combines sample-efficient dynamic programming with maximum likelihood policy updates, providing a simple and effective framework that is able to leverage large amounts of offline data and then quickly perform online fine-tuning of RL policies. We show that our method, advantage weighted actor critic (AWAC), enables rapid learning of skills with a combination of prior demonstration data and online experience. We demonstrate these benefits on a variety of simulated and real-world robotics domains, including dexterous manipulation with a real multi-fingered hand, drawer opening with a robotic arm, and rotating a valve. Our results show that incorporating prior data can reduce the time required to learn a range of robotic skills to practical time-scales.
\end{abstract}

\section{Introduction}\label{sec:introduction}
Learning models that generalize effectively to complex open-world settings, from image recognition~\citep{krizhevsky2012imagenet} to natural language processing~\citep{devlin2019bert}, relies on large, high-capacity models as well as large, diverse, and representative datasets.
Leveraging this recipe of pre-training from large-scale offline datasets has the potential to provide significant benefits for reinforcement learning (RL) as well, both in terms of generalization and sample complexity.
But most existing RL algorithms collect data online from scratch every time a new policy is learned, which can quickly become impractical in domains like robotics where physical data collection has a non-trivial cost. 
In the same way that powerful models in computer vision and NLP are often pre-trained on large, general-purpose datasets and then fine-tuned on task-specific data, practical instantiations of reinforcement learning for real world robotics problems will need to be able to incorporate large amounts of prior data effectively into the learning process, while still collecting additional data online for the task at hand. 
Doing so effectively will make the online data collection process much more practical while still allowing robots operating in the real world to continue improving their behavior. 

\begin{figure}[!t]
    \center
    \includegraphics[width=0.32\linewidth]{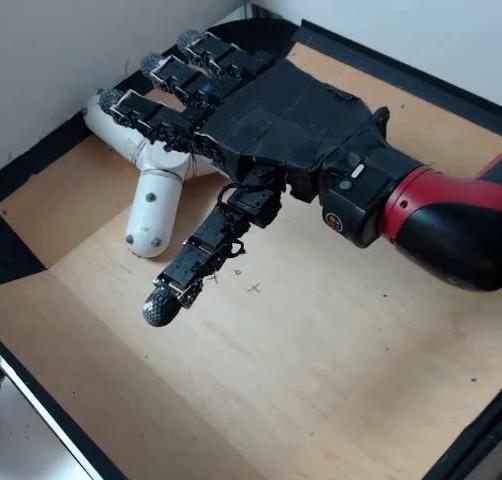}
    \includegraphics[width=0.32\linewidth]{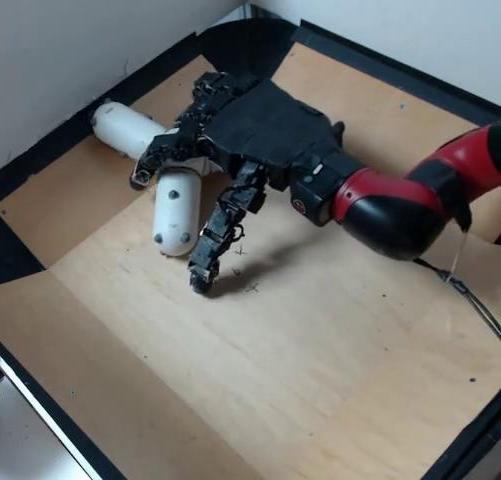}
    \includegraphics[width=0.32\linewidth]{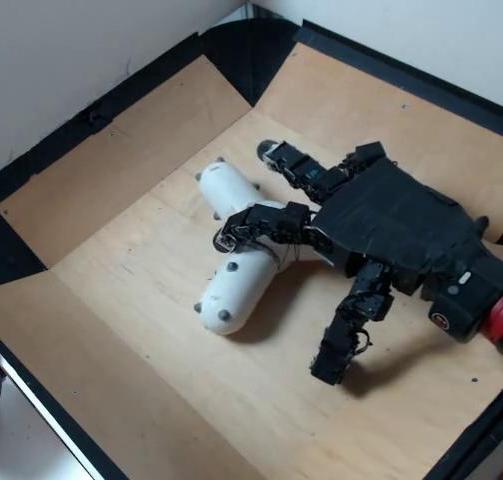}
    \caption{Utilizing prior data for online learning allows us to solve challenging real-world robotics tasks, such as this dexterous manipulation task where the learned policy must control a 4-fingered hand to reposition an object.
    }
    \label{fig:robots}
\end{figure}

For data-driven reinforcement learning, offline datasets consist of trajectories of states, actions and associated rewards. This data can potentially come from demonstrations for the desired task~\citep{schaal97lfd, atkeson1997lfd}, suboptimal policies~\citep{gao2018imperfect}, demonstrations for related tasks~\citep{zhou2019wtl}, or even just random exploration in the environment. Depending on the quality of the data that is provided, useful knowledge can be extracted about the dynamics of the world, about the task being solved, or both. Effective data-driven methods for deep reinforcement learning should be able to use this data to pre-train offline while improving with online fine-tuning. 

\setlength{\tabcolsep}{6pt}
\renewcommand{\arraystretch}{1.5}
\begin{figure*}[!t]
    \includegraphics[width=0.6\textwidth]{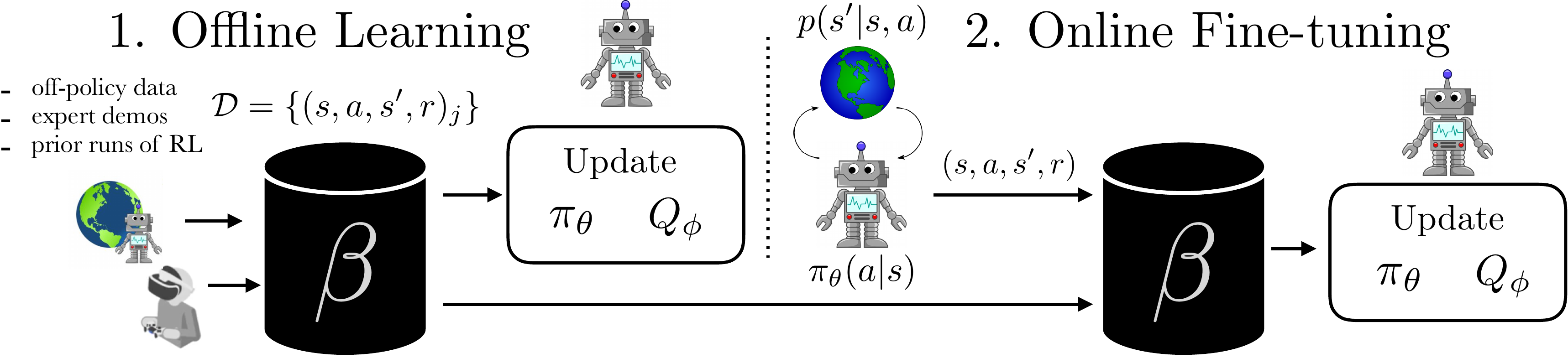}
    \caption{
    We study learning policies by offline learning on a prior dataset $\mathcal{D}$ and then fine-tuning with online interaction. The prior data could be obtained via prior runs of RL, expert demonstrations, or any other source of transitions. Our method, advantage weighted actor critic (AWAC) is able to learn effectively from offline data and fine-tune in order to reach expert-level performance after collecting a limited amount of interaction data.
    Videos and data are available at \projectpage
    \vspace{-0.3cm}
    }
    \label{fig:figure1}
\end{figure*}

Since this prior data can come from a variety of sources, we would like to design an algorithm that does not utilize different types of data in any privileged way. For example, prior methods that incorporate demonstrations into RL directly aim to mimic these demonstrations~\citep{nair2018demonstrations}, which is desirable when the demonstrations are known to be optimal, but imposes strict requirements on the type of offline data, and can cause undesirable bias when the prior data is not optimal. While prior methods for fully offline RL provide a mechanism for utilizing offline data~\citep{fujimoto19bcq, kumar19bear}, as we will show in our experiments, such methods generally are not effective for fine-tuning with online data as they are often too conservative. In effect, prior methods require us to choose: Do we assume prior data is optimal or not? Do we use only offline data, or only online data? To make it feasible to learn policies for open-world settings, we need algorithms that learn successfully in any of these cases. 

In this work, we study how to build RL algorithms that are effective for pre-training from off-policy datasets, but also well suited to continuous improvement with online data collection. We systematically analyze the challenges with using standard off-policy RL algorithms~\citep{haarnoja2018sac, kumar19bear, we2018mpo} for this problem, and introduce a simple actor critic algorithm that elegantly bridges data-driven pre-training from offline data and improvement with online data collection. Our method, which uses dynamic programming to train a critic but a supervised learning style update to train a constrained actor, combines the best of supervised learning and actor-critic algorithms. Dynamic programming can leverage off-policy data and enable sample-efficient learning. The simple supervised actor update implicitly enforces a constraint that mitigates the effects of distribution shift when learning from offline data~\citep{fujimoto19bcq, kumar19bear}, while avoiding overly conservative updates.

We evaluate our algorithm on a wide variety of robotic control tasks, using a set of simulated dexterous manipulation problems as well as three separate real-world robots: drawer opening with a 7-DoF robotic arm, picking up an object with a multi-fingered hand, and rotating a valve with a 3-fingered claw. Our algorithm, \METHOD (AWAC), is able to quickly learn successful policies for these challenging tasks, in spite of high dimensional action spaces and uninformative, sparse reward signals. We show that AWAC finetunes much more efficiently after offline pretraining as compared to prior methods and, given a fixed time budget, attains significantly better performance on the real-world tasks. Moreover, AWAC can utilize different types of prior data without any algorithmic changes: demonstrations, suboptimal data, or random exploration data. The contribution of this work is not just another RL algorithm, but a systematic study of what makes offline pre-training with online fine-tuning unique compared to the standard RL paradigm, which then directly motivates a simple algorithm, AWAC, to address these challenges. We additionally discuss the design decisions required for applying AWAC as a practical tool for real-world robotic skill learning.  

\section{Preliminaries}\label{sec:background}
We use the standard reinforcement learning notation, with states $\st$, actions $\at$, policy $\pi(\at|\st)$, rewards $r(\st, \at)$, and dynamics $p(\st'|\st, \at)$. The discounted return is defined as $R_t = \sum_{i=t}^T \gamma^i r(\st_i, \at_i)$, for a discount factor $\gamma$ and horizon $T$ which may be infinite. The objective of an RL agent is to maximize the expected discounted return $J(\pi) = \E_{p_\pi(\tau)}[R_0]$ under the distribution induced by the policy. 
The optimal policy can be learned directly (e.g., using policy gradient to estimate $\nabla J(\pi)$~\citep{williams1992reinforce}), but this is often ineffective due to high variance of the estimator. Many algorithms attempt to reduce this variance by making use of the value function $V^\pi(\st) = \E_{p_\pi(\tau)} [R_t|\st]$, action-value function $Q^\pi(\st, \at) = \E_{p_\pi(\tau)} [R_t|\st, \at]$, or advantage $A^\pi(\st, \at) = Q^\pi(\st, \at) - V^\pi(\st)$. The action-value function for a policy can be written recursively via the Bellman equation:
\begin{align}
    Q^\pi(\st, \at) &= r(\st, \at) + \gamma \E_{p(\st'|\st,\at)} [V^\pi(\st')] \\ &= r(\st, \at) + \gamma \E_{p(\st'|\st,\at)} [\E_{\pi(\at'|\st')}[Q^\pi(\st', \at')]].
    \label{eq:bellman}
\end{align}
Instead of estimating policy gradients directly, actor-critic algorithms maximize returns by alternating between two phases~\citep{konda2000actorcritic}: policy evaluation and policy improvement. During the policy evaluation phase, the critic $Q^{\pi}(\st,\at)$ is estimated for the current policy $\pi$. This can be accomplished by repeatedly applying the Bellman operator $\mathcal{B}$, corresponding to the right-hand side of Equation~\ref{eq:bellman}, as defined below:
\begin{align} \label{eqn:policy_evaluation}
    \mathcal{B}^\pi Q(\st, \at) & = r(\st, \at) + \gamma \E_{p(\st'|\st,\at)} [\E_{\pi(\at'|\st')}[Q^\pi(\st', \at')]].
\end{align}
By iterating according to $Q^{k+1} = \mathcal{B}^\pi Q^k$, $Q^k$ converges to $Q^\pi$~\citep{sutton1998rl}. With function approximation, we cannot apply the Bellman operator exactly, and instead minimize the Bellman error with respect to Q-function parameters $\phi_k$:
\begin{align} \label{eqn:phi_obj}
    \phi_k &= \arg\min_\phi \; \E_\mathcal{D}[(Q_\phi(\st, \at) - y)^2], \\ y &= r(\st, \at) + \gamma \E_{\st',\at'}[Q_{\phi_{k-1}}(\st', \at')].
\end{align}
During policy improvement, the actor $\pi$ is typically updated based on the current estimate of $Q^\pi$. A commonly used technique~\citep{lillicrap2015continuous, fujimoto2018td3, haarnoja2018sac} is to update the actor $\pi_{\theta_k}(\at|\st)$ via likelihood ratio or pathwise derivatives to optimize the following objective, such that the expected value of the Q-function $Q^\pi$ is maximized:
\begin{align} \label{eqn:policy_improvement}
    \theta_k & = \argmax_{\theta} \E_{\st \sim \mathcal{D}}[\E_{\pi_\theta(\at|\st)}[Q_{\phi_k}(\st,\at)]]
\end{align}
Actor-critic algorithms are widely used in deep RL~\citep{mnih2016asynchronous, lillicrap2015continuous, haarnoja2018sac, fujimoto2018td3}. With a Q-function estimator, they can in principle utilize off-policy data when used with a replay buffer for storing prior transition tuples, which we will denote $\beta$, to sample previous transitions, although we show that this by itself is insufficient for our problem setting. 

\section{Challenges in Offline RL with Online Fine-tuning}
\label{sec:challenges}

In this section, we study the unique challenges that exist when pre-training using offline data, followed by fine-tuning with online data collection. We first describe the problem, and then analyze what makes this problem difficult for prior methods.

\subsection{Problem Definition} \label{sec:challenges_setting}
A static dataset of transitions, \mbox{$\mathcal{D} = \{(\st, \at, \st', r)_j\}$}, is provided to the algorithm at the beginning of training. This dataset can be sampled from an arbitrary policy or mixture of policies, and may even be collected by a human expert. This definition is general and encompasses many scenarios: learning from demonstrations, random data, prior RL experiments, or even from multi-task data. Given the dataset $\mathcal{D}$, our goal is to leverage $\mathcal{D}$ for pre-training and use a small amount of online interaction to learn the optimal policy $\pi^*(\at|\st)$, as depicted in Fig~\ref{fig:figure1}. This setting is representative of many real-world RL settings, where prior data is available and the aim is to learn new skills efficiently. We first study existing algorithms empirically in this setting on the HalfCheetah-v2 Gym environment\footnote{We use this environment for analysis because it helps understand and accentuate the differences between different algorithms. More challenging environments like the ones shown in Fig~\ref{fig:hand-learning-curves}
are too hard to solve to analyze variants of different methods.}. The prior dataset consists of 15 demonstrations from an expert policy and 100 suboptimal trajectories sampled from a behavioral clone of these demonstrations. All methods for the remainder of this paper incorporate the prior dataset, unless explicitly labeled ``scratch''.

\begin{figure*}[t]
    \centering
    \begin{subfigure}[b]{0.01\textwidth}
        \center
        \begin{turn}{90} 
            \scriptsize
            Average Returns
        \end{turn}
        \vspace{0.5cm}
    \end{subfigure}
    \begin{subfigure}[b]{0.27\textwidth}
        \center
        \includegraphics[height=2.2cm]{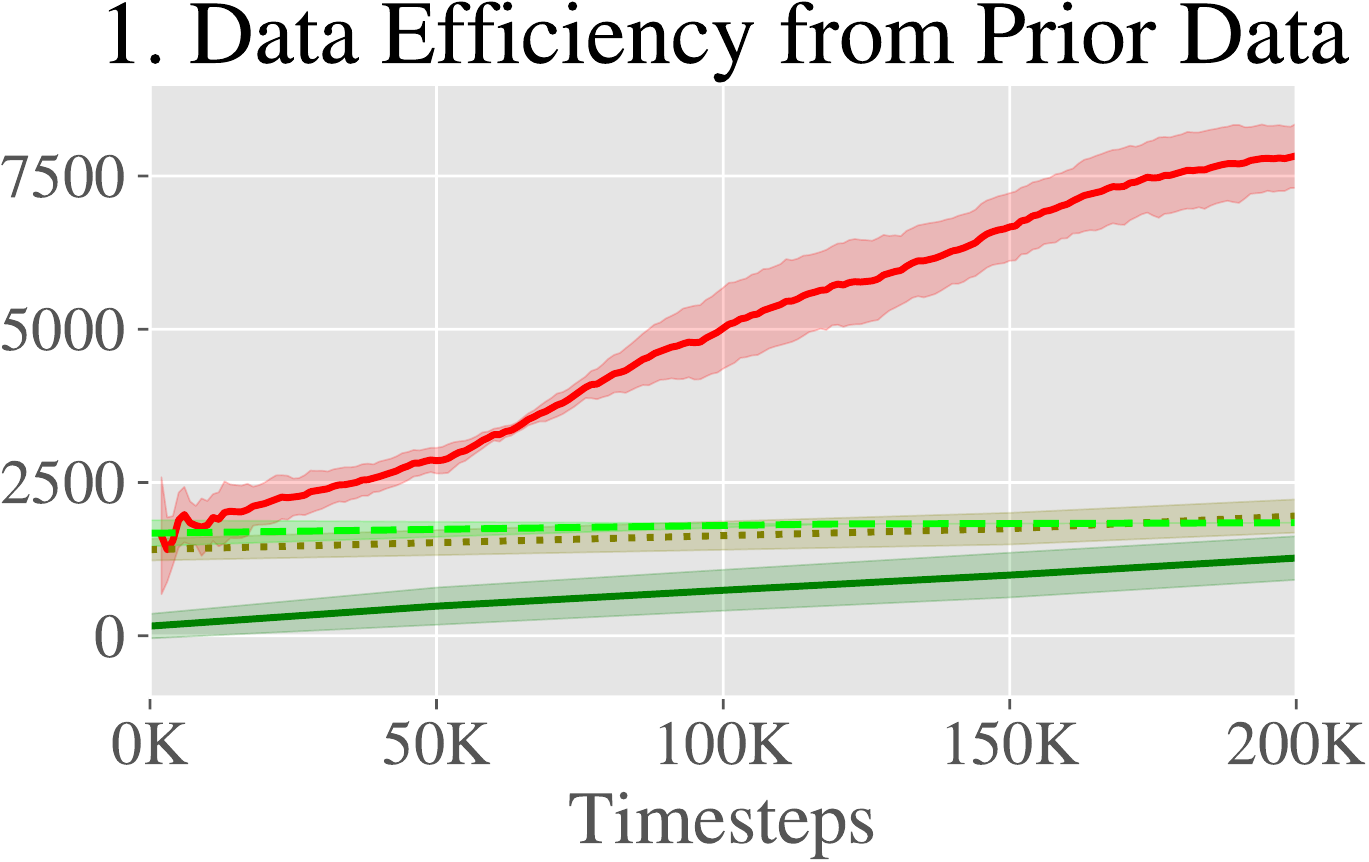}
        \includegraphics[height=0.55cm]{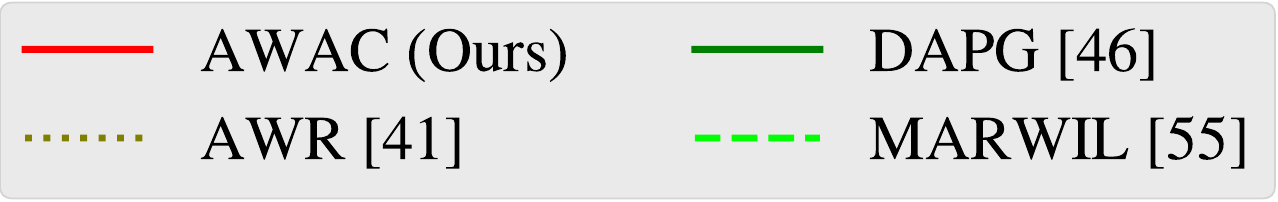}
    \end{subfigure}
    \begin{subfigure}[b]{0.25\textwidth}
        \center
        \includegraphics[height=2.2cm]{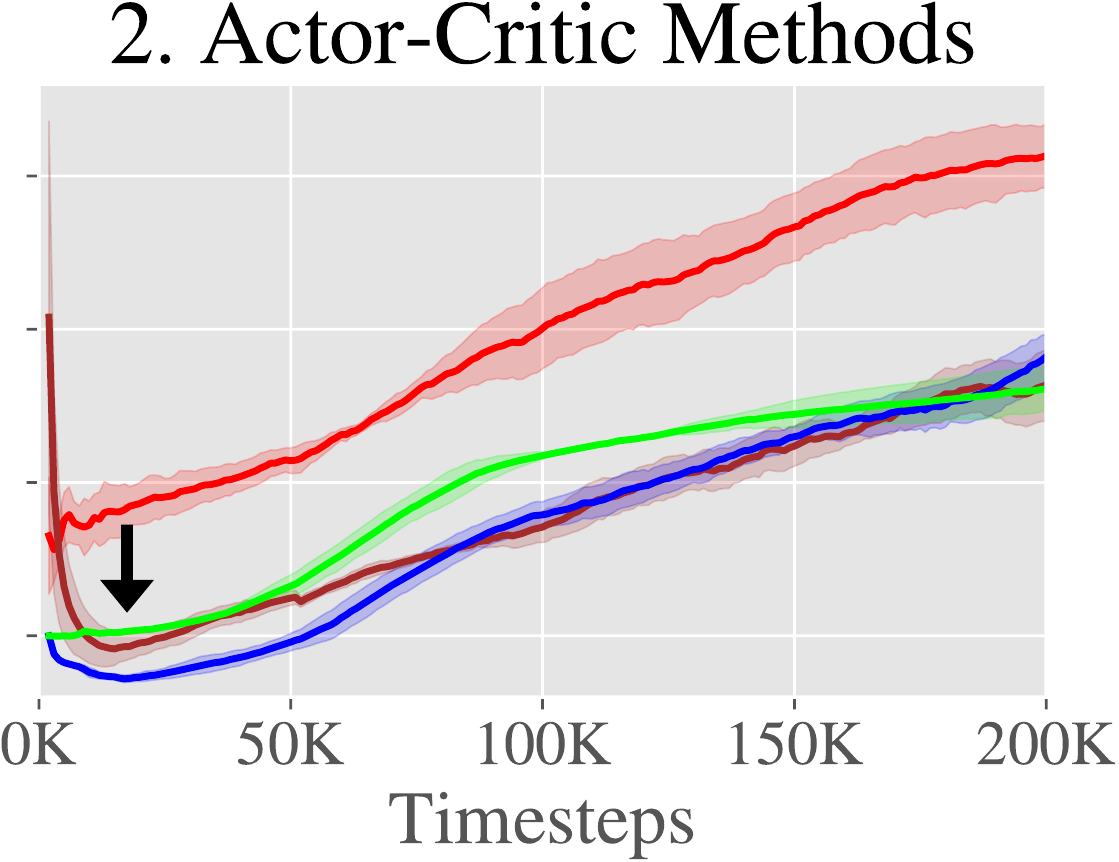}
        \includegraphics[height=0.55cm]{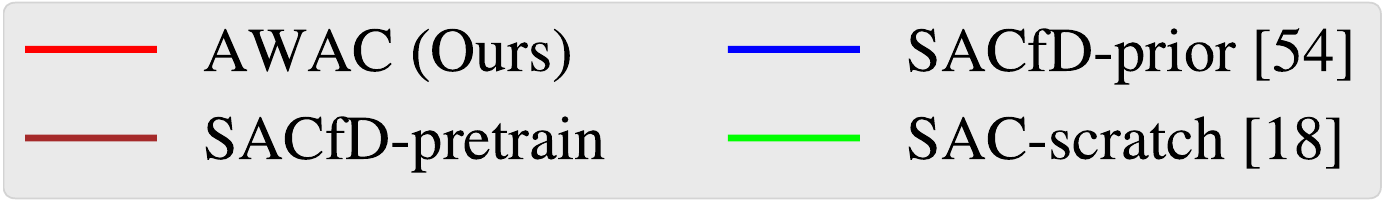}
    \end{subfigure}
    \begin{subfigure}[b]{0.22\textwidth}
        \center
        \includegraphics[height=2.2cm]{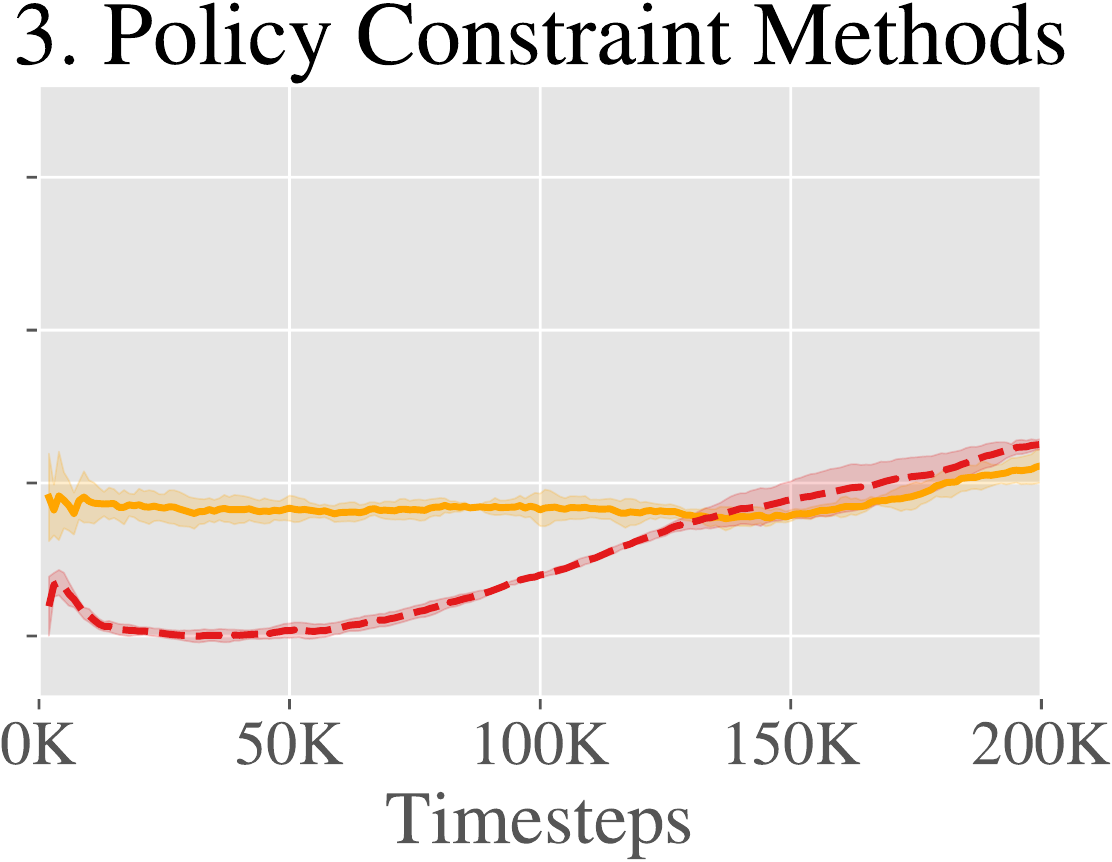}
        \includegraphics[height=0.55cm]{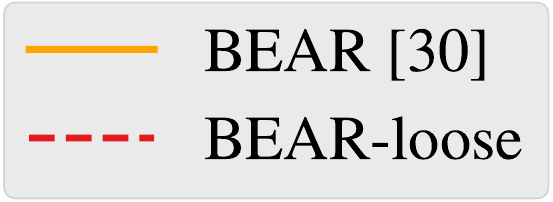}
    \end{subfigure}
    \begin{subfigure}[b]{0.2\textwidth}
        \center
        \includegraphics[height=2.2cm]{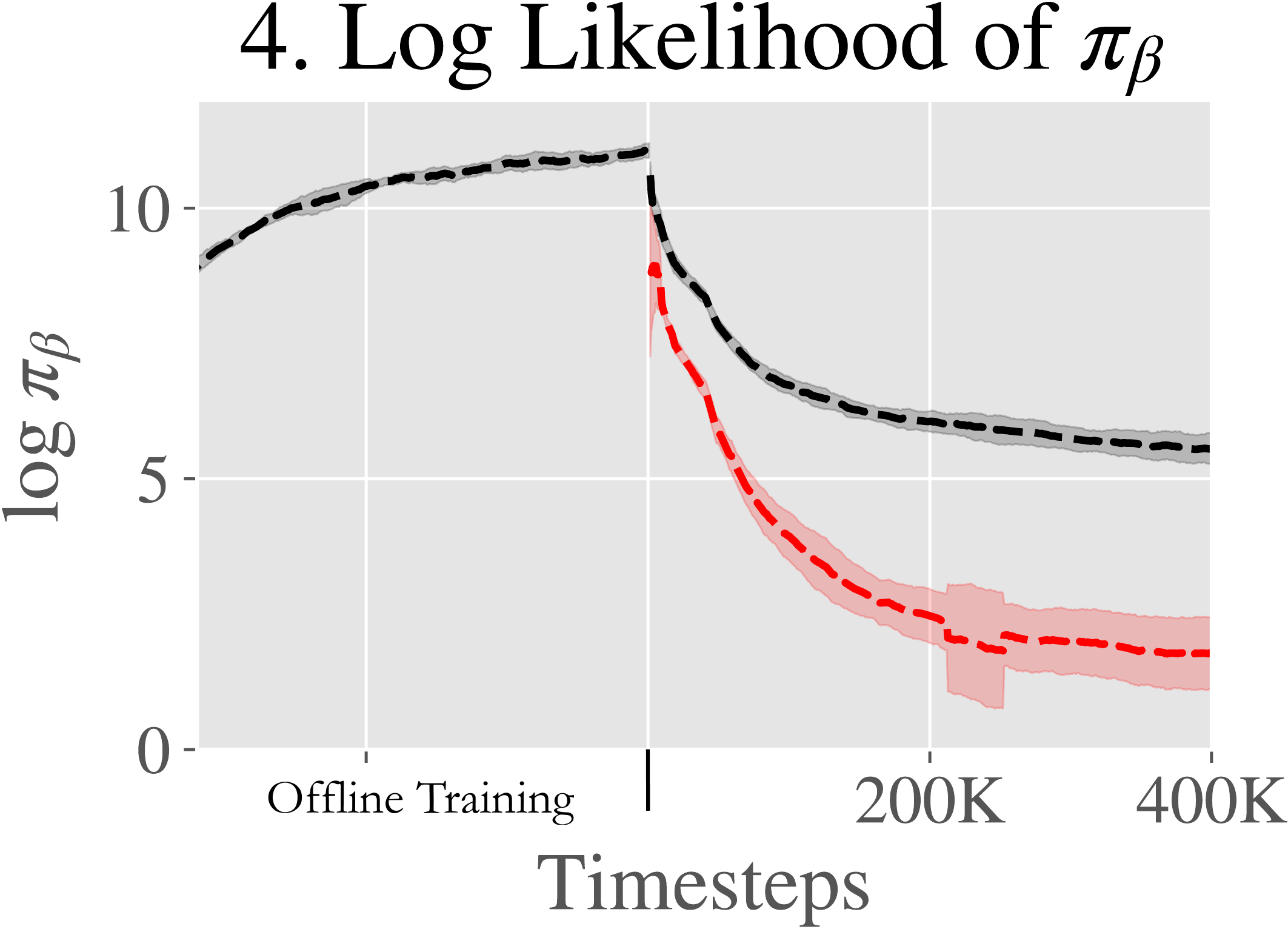}
        \hspace{0.3cm}\includegraphics[height=0.55cm]{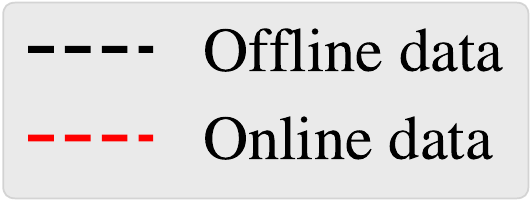}
    \end{subfigure}
    
    \caption{
    Analysis of prior methods on HalfCheetah-v2 using offline RL with online fine-tuning. (1) On-policy methods (DAPG, AWR, MARWIL) 
    learn relatively slowly, even with access to prior data. We present our method, AWAC, as an example of how off-policy RL methods can learn much faster. (2) Variants of soft actor-critic (SAC) with offline training (performed before timestep 0) and fine-tuning. We see a ``dip'' in the initial performance, even if the policy is pretrained with behavioral cloning.
    (3) Offline RL method BEAR~\citep{kumar19bear} on offline training and fine-tuning, including a ``loose'' variant of BEAR with a weakened constraint. Standard offline RL methods fine-tune slowly, while the ``loose'' BEAR variant experiences a similar dip as SAC. (4) We show that the fit of the behavior models $\hat{\pi}_\beta$ used by these offline methods
    degrades as new data is added to the buffer during fine-tuning, potentially explaining their poor fine-tuning performance.
    \vspace{-0.2in}
    }
    \label{fig:challenges}
\end{figure*}

\subsection{Data Efficiency} \label{sec:challenges_efficiency}
One of the simplest ways to utilize prior data such as demonstrations for RL is to pre-train a policy with imitation learning, and fine-tune with on-policy RL~\citep{gupta2019relay, rajeswaran2018dextrous}. This approach has two drawbacks: (1) prior data may not be optimal; (2) on-policy fine-tuning is data inefficient as it does not reuse the prior data in the RL stage. In our setting, data efficiency is vital. To this end, we require algorithms that are able to reuse arbitrary off-policy data during online RL for data-efficient fine-tuning. We find that algorithms that use on-policy fine-tuning~\citep{rajeswaran2018dextrous, gupta2019relay}, or Monte-Carlo return estimation~\citep{peters2007rwr, wang2018marwil, peng2019awr} are generally much less efficient than off-policy actor-critic algorithms, which iterate between improving $\pi$ and estimating $Q^\pi$ via Bellman backups. This can be seen from the results in Figure~\ref{fig:challenges} plot 1, where on-policy methods like DAPG~\citep{rajeswaran2018dextrous} and Monte-Carlo return methods like AWR~\citep{peng2019awr} and MARWIL~\citep{wang2018marwil} are an order of magnitude slower than off-policy actor-critic methods. Actor-critic methods, shown in Figure~\ref{fig:challenges} plot 2, can in principle use off-policy data. However, as we will discuss next, na\"{i}vely applying these algorithms to our problem suffers from a different set of challenges. 

\subsection{Bootstrap Error in Offline Learning with Actor-Critic Methods} \label{sec:challenges_sac}
When standard off-policy actor-critic methods are applied to this problem setting, they perform poorly, as shown in the second plot in Figure~\ref{fig:challenges}: despite having a prior dataset in the replay buffer, these algorithms do not benefit significantly from offline training.
We evaluate soft actor critic~\citep{haarnoja2018sac}, a state-of-the-art actor-critic algorithm for continuous control.
Note that ``SAC-scratch,'' which does not receive the prior data, performs similarly to ``SACfD-prior,'' which does have access to the prior data, indicating that the off-policy RL algorithm is not actually able to make use of the off-policy data for pre-training.
Moreover, even if the SAC is policy is pre-trained by behavior cloning, labeled ``SACfD-pretrain'', we still observe an initial decrease in performance, and performance similar to learning from scratch.

This challenge can be attributed to off-policy bootstrapping error accumulation, as observed in several prior works~\citep{sutton1998rl, kumar19bear, wu2019brac, levine2020offlinetutorial, fujimoto19bcq}.
In actor-critic algorithms, the target value $Q(\st', \at')$, with $\at' \sim \pi$, is used to update $Q(\st, \at)$. When $\at'$ is outside of the data distribution, $Q(\st', \at')$ will be inaccurate, leading to accumulation of error on static datasets. 

Offline RL algorithms~\citep{fujimoto19bcq, kumar19bear, wu2019brac} propose to address this issue by explicitly adding constraints on the policy improvement update (Equation~\ref{eqn:policy_improvement}) to avoid bootstrapping on out-of-distribution actions, leading to a policy update of this form:
\begin{align} \label{eqn:offline_pi_constraint}
    \argmax_{\theta} \E_{\st \sim \mathcal{D}}[\E_{\pi_\theta(\at|\st)}[Q_{\phi_k}(\st,\at)]] \; \text{s.t.} \; D(\pi_\theta, \pi_\beta) \leq \epsilon.
\end{align}
Here, $\pi_\theta$ is the actor being updated, and $\pi_\beta(a|s)$ represents the (potentially unknown) distribution from which all of the data seen so far (both offline data and online data) was generated. In the case of a replay buffer, $\pi_\beta$ corresponds to a mixture distribution over all past policies. Typically, $\pi_\beta$ is not known, especially for offline data, and must be estimated from the data itself. Many offline RL algorithms~\citep{kumar19bear, fujimoto19bcq, siegel2020abm} explicitly fit a parametric model to samples for the distribution $\pi_\beta$ via maximum likelihood estimation, where samples from $\pi_\beta$ are obtained simply by sampling uniformly from the data seen thus far:
$
    \hat{\pi}_\beta = \max_{\hat{\pi}_\beta} \; \E_{\st, \at \sim \pi_\beta}[\log \hat{\pi}_\beta(\at|\st)]
$.
After estimating $\hat{\pi}_\beta$, prior methods implement the constraint given in Equation~\ref{eqn:offline_pi_constraint} in various ways, including penalties on the policy update~\citep{kumar19bear,wu2019brac} or architecture choices for sampling actions for policy training~\citep{fujimoto19bcq,siegel2020abm}. As we will see next, the requirement for accurate estimation of $\hat{\pi}_\beta$ makes these methods difficult to use with online fine-tuning.

\subsection{Excessively Conservative Online Learning} \label{sec:challenges_bear}
While offline RL algorithms with constraints~\citep{kumar19bear, fujimoto19bcq, wu2019brac} perform well offline, they struggle to improve with fine-tuning, as shown in the third plot in Figure~\ref{fig:challenges}. We see that the purely offline RL performance (at ``0K'' in Fig.~\ref{fig:challenges}) is much better than the standard off-policy methods shown in Section~\ref{sec:challenges_sac}. However, with additional iterations of online fine-tuning, the performance increases very slowly (as seen from the slope of the BEAR curve in Fig~\ref{fig:challenges}). What causes this phenomenon?

This can be attributed to challenges in fitting an accurate behavior model as data is collected online during fine-tuning. In the offline setting, behavior models must only be trained once via maximum likelihood, but in the online setting, the behavior model must be updated online to track incoming data. Training density models online (in the ``streaming'' setting) is a challenging research problem~\citep{ramapuram2017lifelonggm}, made more difficult by a potentially complex multi-modal behavior distribution induced by the mixture of online and offline data. To understand this, we plot the log likelihood of learned behavior models on the dataset during online and offline training for the HalfCheetah task. As we can see in the plot, the accuracy of the behavior models ($\log \pi_\beta$ on the y-axis) reduces during online fine-tuning, indicating that it is not fitting the new data well during online training. When the behavior models are inaccurate or unable to model new data well, constrained optimization becomes too conservative, resulting in limited improvement with fine-tuning. This analysis suggests that, in order to address our problem setting, we require an off-policy RL algorithm that constrains the policy to prevent offline instability and error accumulation, but not so conservatively that it prevents online fine-tuning due to imperfect behavior modeling. Our proposed algorithm, which we discuss in the next section, accomplishes this by employing an \emph{implicit} constraint, which does not require \emph{any} explicit modeling of the behavior policy.

\section{Advantage Weighted Actor Critic: A Simple Algorithm for Fine-tuning from Offline Datasets}
\label{sec:method}

In this section, we will describe the advantage weighted actor-critic (AWAC) algorithm, which trains an off-policy critic and an actor with an \emph{implicit} policy constraint. We will show AWAC mitigates the challenges outlined in Section~\ref{sec:challenges}. AWAC follows the design for actor-critic algorithms as described in Section~\ref{sec:background}, with a policy evaluation step to learn $Q^\pi$ and a policy improvement step to update $\pi$. AWAC uses off-policy temporal-difference learning to estimate $Q^\pi$ in the policy evaluation step, and a policy improvement update that is able to obtain the benefits of offline RL algorithms at training from prior datasets, while avoiding the overly conservative behavior described in Section~\ref{sec:challenges_bear}. We describe the policy improvement step in AWAC below, and then summarize the entire algorithm. 

Policy improvement for AWAC proceeds by learning a policy that maximizes the value of the critic learned in the policy evaluation step via TD bootstrapping. If done naively, this can lead to the issues described in Section~\ref{sec:challenges_bear}, but we can avoid the challenges of bootstrap error accumulation by restricting the policy distribution to stay close to the data observed thus far during the actor update, while maximizing the value of the critic. At iteration $k$, AWAC therefore optimizes the policy to maximize the estimated Q-function $Q^{\pi_k}(\st, \at)$ at every state, while constraining it to stay close to the actions observed in the data, similar to prior offline RL methods, though this constraint will be enforced differently. Note from the definition of the advantage in Section~\ref{sec:background} that optimizing $Q^{\pi_k}(\st, \at)$ is equivalent to optimizing $A^{\pi_k}(\st, \at)$. We can therefore write this optimization as:
\begin{align} \label{eqn:pi_obj}
    \pinew = \; \argmax_{\pi \in \Pi} \;  \E_{\at \sim \pi(\cdot|\st)}[A^{\pi_k}(\st, \at)] \\
    \text{s.t.} \; \KL(\pi(\cdot|\st)||\pi_\buffer(\cdot|\st)) \leq \epsilon.
\end{align} 
As we saw in Section~\ref{sec:challenges_sac}, enforcing the constraint by incorporating an explicit learned behavior model~\citep{kumar19bear, fujimoto19bcq, wu2019brac, siegel2020abm} leads to poor fine-tuning performance. Instead, we enforce the constraint \emph{implicitly}, without learning a behavior model. We first derive the solution to the constrained optimization in Equation~\ref{eqn:pi_obj} to obtain a non-parametric closed form for the actor. This solution is then projected onto the parametric policy class \emph{without} any explicit behavior model. 
The analytic solution to Equation \ref{eqn:pi_obj} can be obtained by enforcing the KKT conditions~\citep{peters2007rwr, peters2010reps, peng2019awr}. The Lagrangian is:
\begin{align}
    \mathcal{L}(\pi, \lagrangeawr) = \E_{\at \sim \pi(\cdot|\st)}[A^{\pi_k}(\st, \at)] + \lagrangeawr(\epsilon - \KL(\pi(\cdot|\st)||\piold(\cdot|\st))),
\end{align}
and the closed form solution to this problem is
$\pi^*(\at|\st) \propto \pi_\buffer(\at|\st)\exp\left(\frac{1}{\lagrangeawr} A^{\pi_k}(\st, \at)\right)$.
When using function approximators, such as deep neural networks as we do, we need to project the non-parametric solution into our policy space. For a policy $\pi_\theta$ with parameters $\theta$, this can be done by minimizing the KL divergence of $\pi_{\theta}$ from the optimal non-parametric solution $\pi^*$ under the data distribution $\rho_{\pi_\buffer}(\st)$: 
\begin{align}
    & \argmin_\theta \; \Ex_{\rho_{\pi_\buffer}(\st) } \left[\KL(\pi^*(\cdot|\st)||\pi_\theta(\cdot|\st))\right] \\ =
    & \argmin_\theta \; \Ex_{\rho_{\pi_\buffer}(\st)} \left[\Ex_{\pi^*(\cdot|\st)}[-\log \pi_\theta(\cdot|\st)]\right]
\end{align}
Note that the parametric policy could be projected with either direction of KL divergence. Choosing the reverse KL results in explicit penalty methods~\citep{wu2019brac} that rely on evaluating the density of a learned behavior model. 
Instead, by using forward KL, we can compute the policy update by sampling directly from $\beta$:
\begin{align}
    \theta_{k+1} = \argmax_\theta \; & \; \Ex_{\st, \at \sim \buffer}
    \left[\log \pi_\theta(\at|\st) \exp \left(\frac{1}{\lagrangeawr}A^{\pi_k}(\st, \at) \right)\right]. \label{eqn:theta_obj}
\end{align}
This actor update amounts to weighted maximum likelihood (i.e., supervised learning), where the targets are obtained by re-weighting the state-action pairs observed in the current dataset by the predicted advantages from the learned critic, \emph{without} explicitly learning any parametric behavior model, simply sampling $(s, a)$ from the replay buffer $\beta$. See Appendix~\ref{sec:derivation} for a more detailed derivation and Appendix~\ref{sec:implementation} for specific implementation details.

\noindent \textbf{Avoiding explicit behavior modeling.} Note that the update in Equation~\ref{eqn:theta_obj} completely avoids any modeling of the previously observed data $\beta$ with a parametric model. By avoiding any explicit learning of the behavior model AWAC is far less conservative than methods which fit a model $\hat{\pi}_\beta$ explicitly, and better incorporates new data during online fine-tuning, as seen from our results in Section~\ref{sec:experiments}. This derivation is related to AWR~\citep{peng2019awr}, with the main difference that AWAC uses an off-policy Q-function $Q^\pi$ to estimate the advantage, which greatly improves efficiency and even final performance (see results in Section~\ref{sec:dextrous_exps}). The update also resembles ABM-MPO, but ABM-MPO \emph{does} require modeling the behavior policy which, as discussed in Section~\ref{sec:challenges_bear}, can lead to poor fine-tuning. In Section~\ref{sec:dextrous_exps}, AWAC outperforms ABM-MPO on a range of challenging tasks.

\noindent \textbf{Policy evaluation.} During policy evaluation, we estimate the action-value $Q^{\pi}(\st,\at)$ for the current policy $\pi$, as described in Section~\ref{sec:background}. We utilize a temporal difference learning scheme for policy evaluation~\citep{haarnoja2018sac, fujimoto2018td3}, minimizing the Bellman error as described in Equation~\ref{eqn:policy_evaluation}. This enables us to learn very efficiently from off-policy data. This is particularly important in our problem setting to effectively use the offline dataset, and allows us to significantly outperform alternatives using Monte-Carlo evaluation or TD($\lambda$) to estimate returns~\citep{peng2019awr}.  

\begin{algorithm}[H]
  	\caption{Advantage Weighted Actor Critic (AWAC)}
  	\label{alg:method}
  	\begin{algorithmic}[1]
  	\STATE Dataset $\mathcal{D} = \{(\st, \at, \st', r)_j\}$
  	\STATE Initialize buffer $\beta=\mathcal{D}$
  	\STATE Initialize $\pi_\theta$, $Q_\phi$
  	\FOR{iteration $i=1, 2, ...$}
  	    \STATE Sample batch $(\st, \at, \st', r) \sim \beta$
        \STATE $y = r(\st, \at) + \gamma \E_{\st',\at'}[Q_{\phi_{k-1}}(\st', \at')]$
        \STATE $\phi \gets \arg\min_\phi \; \E_\mathcal{D}[(Q_\phi(\st, \at) - y)^2]$
        \STATE $\theta \gets \argmax_\theta \; \Ex_{\st, \at \sim \buffer}
    \left[\log \pi_\theta(\at|\st) \exp \left(\frac{1}{\lagrangeawr}A^{\pi_k}(\st, \at) \right)\right]$
        \IF {$i >$ num\_offline\_steps}
            \STATE $\tau_1, \dots, \tau_K \sim p_{\pi_\theta}(\tau)$
            \STATE $\beta \gets \beta \cup \{\tau_1, \dots, \tau_K\}$
        \ENDIF
  	\ENDFOR
  	\end{algorithmic}
\end{algorithm}

\noindent \textbf{Algorithm summary.} The full AWAC algorithm for offline RL with online fine-tuning is summarized in Algorithm \ref{alg:method}. In a practical implementation, we can parameterize the actor and the critic by neural networks and perform SGD updates from Eqn.~\ref{eqn:theta_obj} and Eqn.~\ref{eqn:phi_obj}. Specific details are provided in Appendix~\ref{sec:implementation}. AWAC ensures data efficiency with off-policy critic estimation via bootstrapping, and avoids offline bootstrap error with a constrained actor update. By avoiding explicit modeling of the behavior policy, AWAC avoids overly conservative updates.

\begin{figure*}[t]
    \center
    \hspace{0.04\textwidth}
    
    
    \centering
    \begin{subfigure}[b]{0.02\textwidth}
        \center
        \begin{turn}{90} 
            \footnotesize
            Success Rate
        \end{turn}
        \vspace{1cm}
    \end{subfigure}
    \begin{subfigure}[b]{0.3\textwidth}
        \center
        \includegraphics[width=1\textwidth]{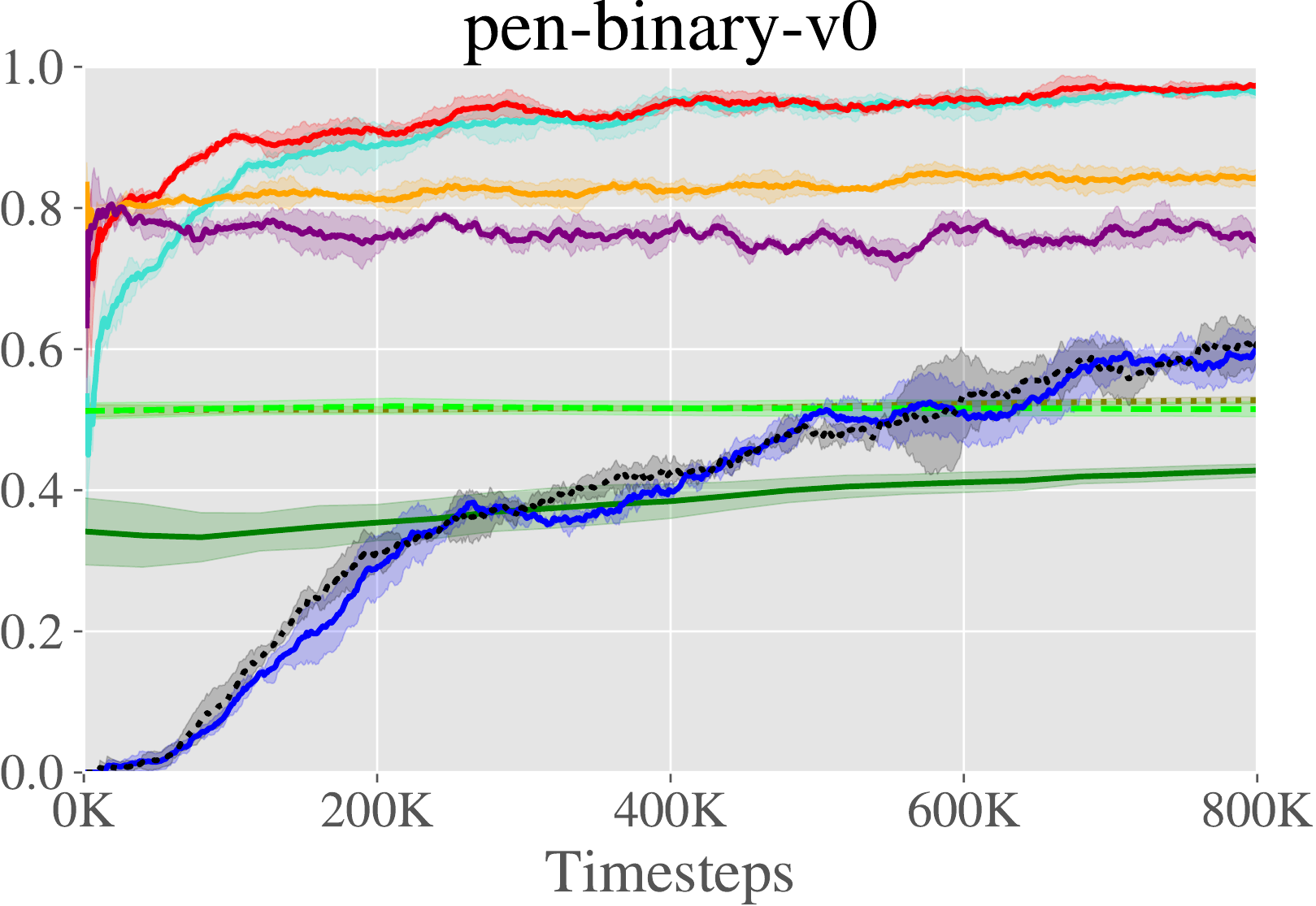}
    \end{subfigure}
    \begin{subfigure}[b]{0.3\textwidth}
        \center
        \includegraphics[width=1\textwidth]{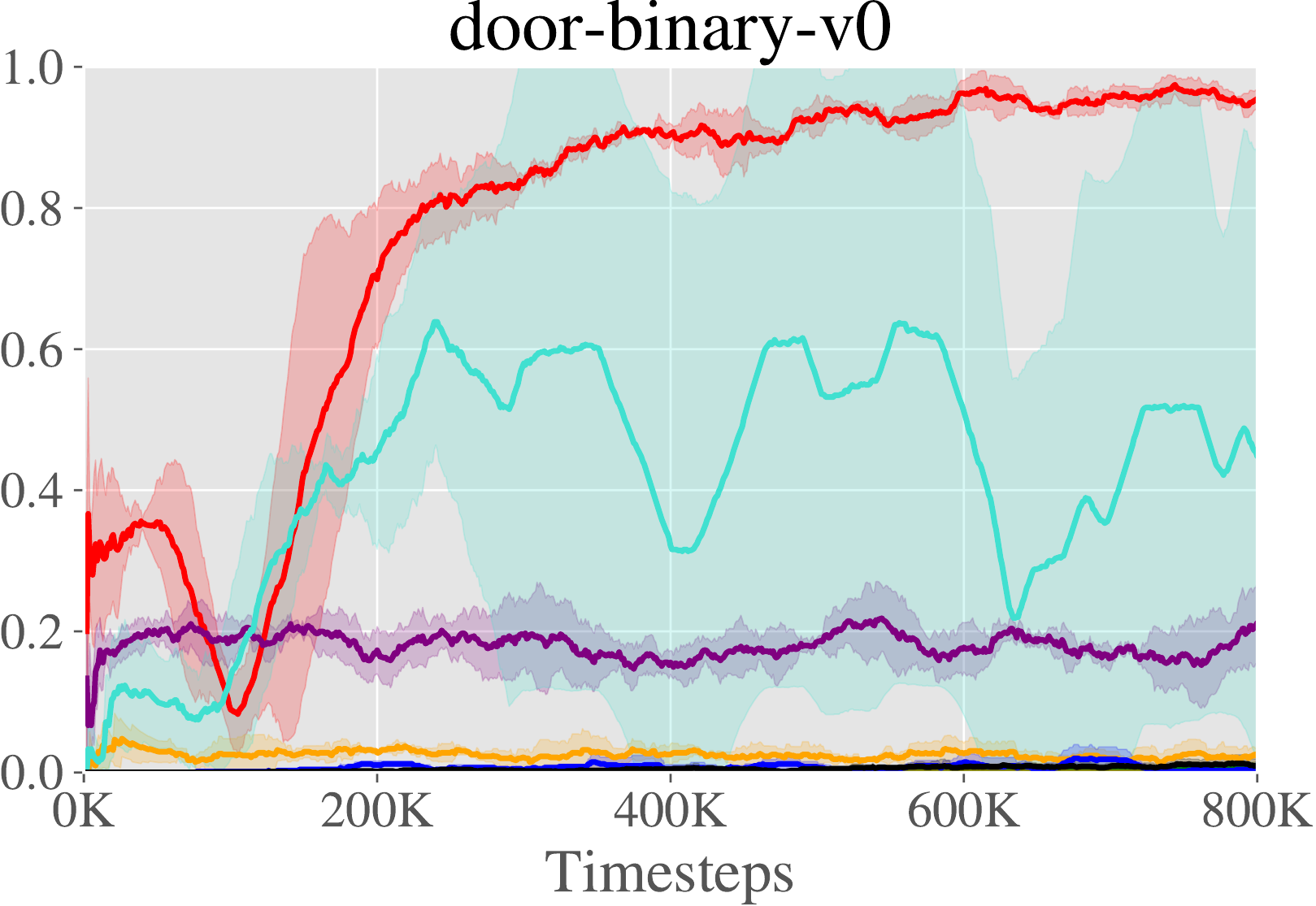}
    \end{subfigure}
    \begin{subfigure}[b]{0.3\textwidth}
        \center
        \includegraphics[width=1\textwidth]{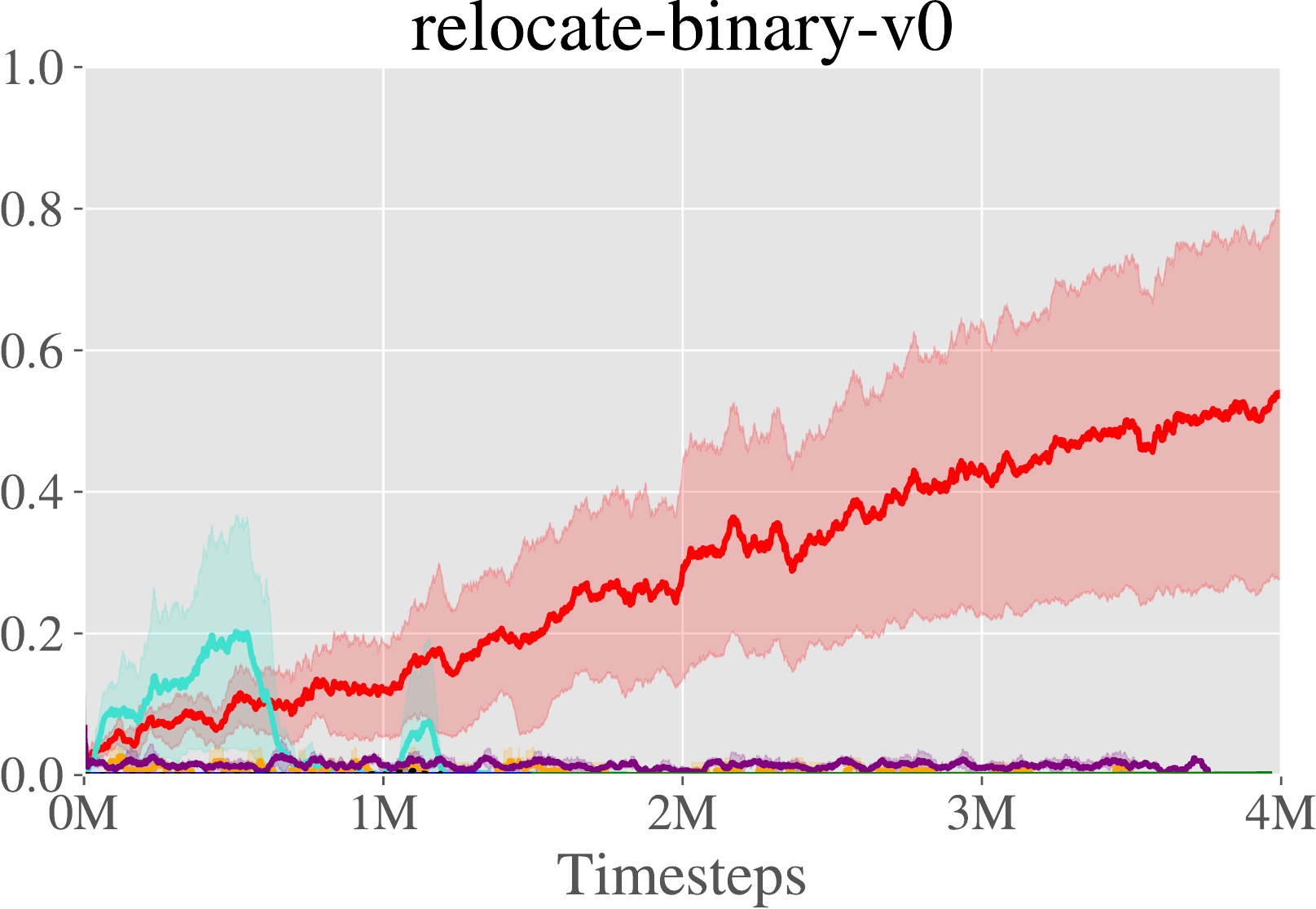}
    \end{subfigure}
    
    \begin{subfigure}[b]{0.99\textwidth}
        \center
        \includegraphics[width=0.6\textwidth]{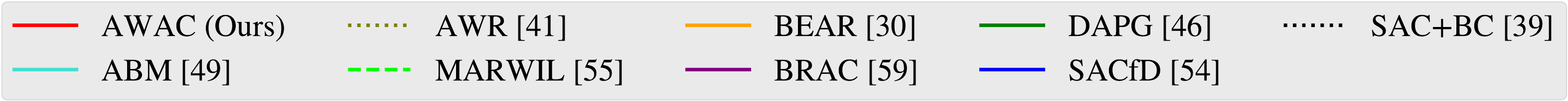}
    \end{subfigure}

    \caption{
    Comparative evaluation on the dexterous manipulation tasks. These tasks are difficult due to their high action dimensionality and reward sparsity. We see that AWAC is able to learn these tasks with little online data collection required (100K samples $\approx$ 16 minutes of equivalent real-world interaction time). Meanwhile, most prior methods are not able to solve the harder two tasks: door opening and object relocation. 
    \vspace{-0.5cm}
    }
    \label{fig:hand-learning-curves}
\end{figure*}

While AWAC is related to several prior RL algorithms, we note that there are key differences that make it particularly amenable to the problem setting we are considering -- offline RL with online fine-tuning -- that other methods are unable to tackle.
As we show in our experimental analysis with direct comparisons to prior work, every one of the design decisions being made in this work are important for algorithm performance. As compared to AWR~\citep{peng2019awr}, AWAC uses TD bootstrapping for significantly more efficient and even asymptotically better performance. As compared to offline RL techniques like ABM~\citep{siegel2020abm}, MPO~\citep{we2018mpo}, BEAR ~\citep{kumar19bear} or BCQ~\citep{fujimoto19bcq} this work is able to avoid the need for any behavior modeling, thereby enabling the \emph{online} fine-tuning part of the problem much better. As shown in Fig~\ref{fig:hand-learning-curves}, when these seemingly ablations are made to AWAC, the algorithm performs significantly worse.

\section{Related Work}\label{sec:related_work}
Off-policy RL algorithms are designed to reuse off-policy data during training, and have been studied extensively~\citep{konda2000actorcritic, degris2012, mnih2016asynchronous, haarnoja2018sac, fujimoto2018td3, bhatnagar2009, peters2008, zhang2019, pawel2009, balduzzi2015}. While standard off-policy methods are able to benefit from including data seen \emph{during} a training run, as we show in Section~\ref{sec:challenges_sac} they struggle when training from previously collected offline data from other policies, due to error accumulation with distribution shift~\citep{fujimoto19bcq, kumar19bear}. Offline RL methods aim to address this issue, often by constraining the actor updates to avoid excessive deviation from the data distribution~\citep{lange2012, thomas2016, hallak2015offpolicy, hallak2016td, hallak2017onlineoffpolicy, agarwal2019optimism, kumar19bear, fujimoto19bcq, rasool2019p3o, nachum2019dualdice, siegel2020abm, levine2020offlinetutorial, zhang2020gendice}. One class of these methods utilize importance sampling~\citep{thomas2016, zhang2020gendice, nachum2019dualdice, degris2012, jiang2016doublyrobust, hallak2017onlineoffpolicy}. 
Another class of methods perform offline reinforcement learning via dynamic programming, with an explicit constraint to prevent deviation from the data distribution ~\citep{lange2012, kumar19bear, fujimoto19bcq, wu2019brac, jaques2019}. While these algorithms perform well in the purely offline settings, we show in Section~\ref{sec:challenges_bear} that such methods tend to be overly conservative, and therefore may not learn efficiently when fine-tuning with online data collection. In contrast, our algorithm AWAC is comparable to these algorithms for offline pre-training, but learns much more efficiently during subsequent fine-tuning. 

Prior work has also considered the special case of learning from \emph{demonstration} data. One class of algorithms initializes the policy via behavioral cloning from demonstrations, and then fine-tunes with reinforcement learning~\citep{peters2008baseball, ijspeert2002attractor, Theodorou2010, kim2013apid, rajeswaran2018dextrous, gupta2019relay, zhu2019hands}. Most such methods use on-policy fine-tuning, which is less sample-efficient than off-policy methods that perform value function estimation. 
Other prior works have incorporated demonstration data into the replay buffer using off-policy RL methods~\citep{vecerik17ddpgfd, nair2017icra}. We show in Section~\ref{sec:challenges_sac} that these strategies can result in a large dip in performance during online fine-tuning, due to the inability to pre-train an effective value function from offline data.
In contrast, our work shows that using supervised learning style policy updates can allow for better bootstrapping from demonstrations as compared to \citet{vecerik17ddpgfd} and \citet{nair2017icra}.

Our method builds on algorithms that implement a maximum likelihood objective for the actor, based on an expectation-maximization formulation of RL~\citep{peters2007rwr,neumann2008fqiawr,Theodorou2010,peters2010reps,peng2019awr,we2018mpo,wang2018marwil}. Most closely related to our method in this respect are the algorithms proposed by~\citet{peng2019awr} (AWR) and \citet{siegel2020abm} (ABM). Unlike AWR, which estimates the value function of the \emph{behavior} policy, $V^{\pi_\beta}$ via Monte-Carlo estimation or TD$-\lambda$, our algorithm estimates the Q-function of the \emph{current} policy $Q^\pi$ via bootstrapping, enabling much more efficient learning, as shown in our experiments. Unlike ABM, our method does not require learning a separate function approximator to model the behavior policy $\pi_\beta$, and instead directly samples the dataset. As we discussed in Section~\ref{sec:challenges_bear}, modeling $\pi_\beta$ can be a major challenge for online fine-tuning. While these distinctions may seem somewhat subtle, they are important and we show in our experiments that they result in a large difference in algorithm performance. Finally, our work goes beyond the analysis in prior work, by studying the issues associated with pre-training and fine-tuning in Section~\ref{sec:challenges}.
Closely to our work, \citet{wang2020crr} proposed critic regularized regression for offline RL, which uses off-policy Q-learning and an equivalent policy update.
In contrast to this concurrent work, we specifically study the offline pretraining online fine-tuning problem, which this prior work does not address, analyze why other methods are ineffective in this setting, and show that our approach enables strong fine-tuning results on challenging dextrous manipulation tasks and real-world robotic systems.

The idea of bootstrapping learning from prior data for real-world robotic learning is not a new one; in fact, it has been extensively explored in the context of providing initial rollouts to bootstrap policy search~\cite{kober2008mp, peters2008baseball, kormushev2010}, initializing dynamic motion primitives~\cite{bentivagna2004, kormushev2010, mulling2013} in the context of on-policy reinforcement learning algorithms~\cite{rajeswaran2018dextrous, zhu2019hands}, inferring reward shaping~\cite{wu2020shaping} and even for inferring reward functions~\cite{ziebart2008maxent, abbeel2004apprenticeship}.
Our work shows how we can generalize the idea of bootstrapping robotic learning from prior data to include arbitrary sub-optimal data rather than just demonstration data and shows the ability to continue improving beyond this data as well. 

\section{Experimental Evaluation}\label{sec:experiments}
In our experimental evaluation we aim to answer the following question: 

\begin{enumerate}
    \item Does AWAC effectively combine prior data with online experience to learn complex robotic control tasks more efficiently than prior methods?
    \item Is AWAC able to learn from sub-optimal or random data?
    \item Does AWAC provide a practical way to bootstrap real-world robotic reinforcement learning?
\end{enumerate}
\noindent In the following sections, we study these questions using several challenging and high-dimensional simulated robotic tasks, as well as three separate real-world robotic platforms.
Videos of all experiments are available at \projectpage

\begin{figure}[tpb]
    \center
    \hspace{0.04\textwidth}
    \begin{subfigure}[b]{0.3\textwidth}
        \center
        \includegraphics[height=0.8\linewidth]{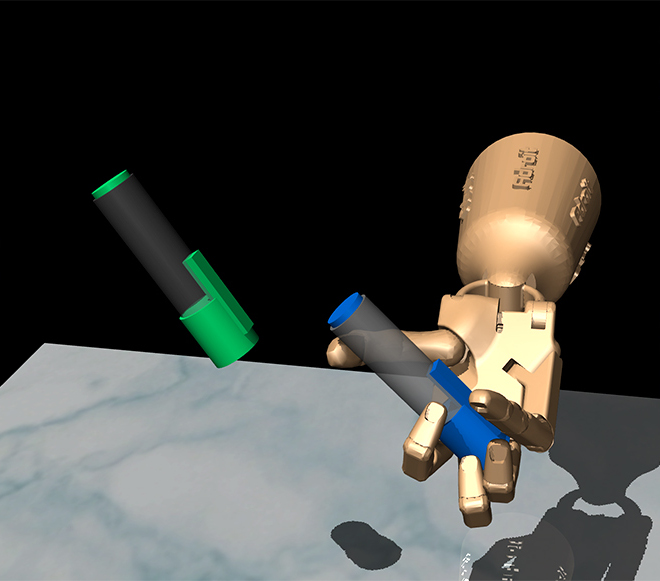}
    \end{subfigure}
    \begin{subfigure}[b]{0.3\textwidth}
        \center
        \includegraphics[height=0.8\linewidth]{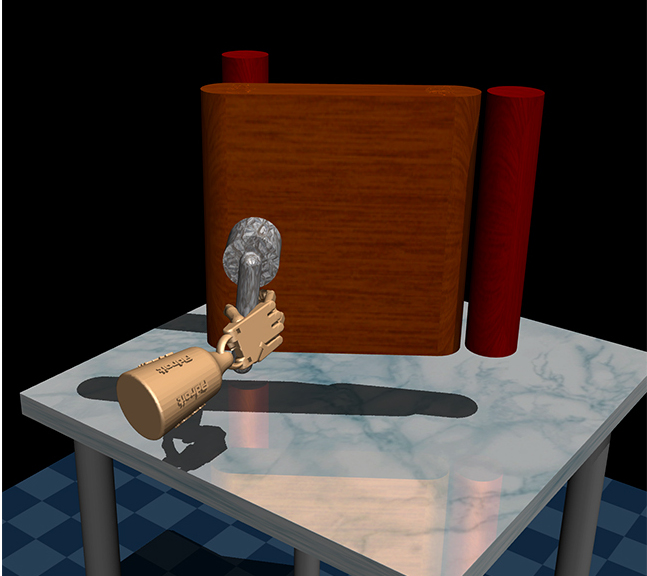}
    \end{subfigure}
    \begin{subfigure}[b]{0.3\textwidth}
        \center
        \includegraphics[height=0.8\linewidth]{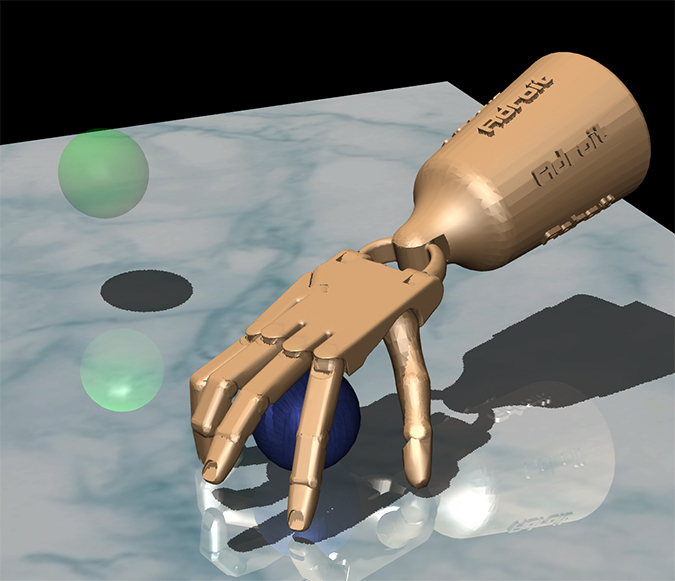}
    \end{subfigure}
    \caption{Illustration of dexterous manipulation tasks in simulation. These tasks exhibit sparse rewards, high-dimensional control, and complex contact physics. }
\end{figure}

\subsection{Simulated Experiments}
We study the first two questions in challenging simulation  environments. 

\subsubsection{Comparative Evaluation When Bootstrapping From Prior Data} \label{sec:dextrous_exps}

We study tasks in simulation that have significant exploration challenges, where offline learning and online fine-tuning are likely to be effective. We begin our analysis with a set of challenging sparse reward dexterous manipulation tasks proposed by \citet{rajeswaran2018dextrous} in simulation.
These tasks involve complex manipulation skills using a 28-DoF five-fingered hand in the MuJoCo simulator~\citep{todorov12mujoco} shown in Figure~\ref{fig:hand-learning-curves}: in-hand rotation of a pen, opening a door by unlatching the handle, and picking up a sphere and relocating it to a target location.
The reward functions in these environments are binary 0-1 rewards for task completion. 
\footnote{\citet{rajeswaran2018dextrous} use a combination of task completion factors as the sparse reward. For instance, in the door task, the sparse reward as a function of the door position $d$ was $r = 10\mathds{1}_{d > 1.35} + 8\mathds{1}_{d > 1.0} + 2\mathds{1}_{d > 1.2} - 0.1||d - 1.57||_2$. We only use the fully sparse success measure $r = \mathds{1}_{d > 1.4}$, which is substantially more difficult. } 
\citet{rajeswaran2018dextrous} provide 25 human demonstrations for each task, which are not fully optimal but do solve the task. Since this dataset is small, we generated another 500 trajectories of interaction data by constructing a behavioral cloned policy, and then sampling from this policy.

\begin{figure*}[t]
    
    \centering
    \begin{subfigure}[b]{0.02\textwidth}
        \center
        \begin{turn}{90} 
            \footnotesize
            Average Return
        \end{turn}
        \vspace{0.8cm}
    \end{subfigure}
    \begin{subfigure}[b]{0.28\textwidth}
        \center
        
        \includegraphics[height=0.22\linewidth]{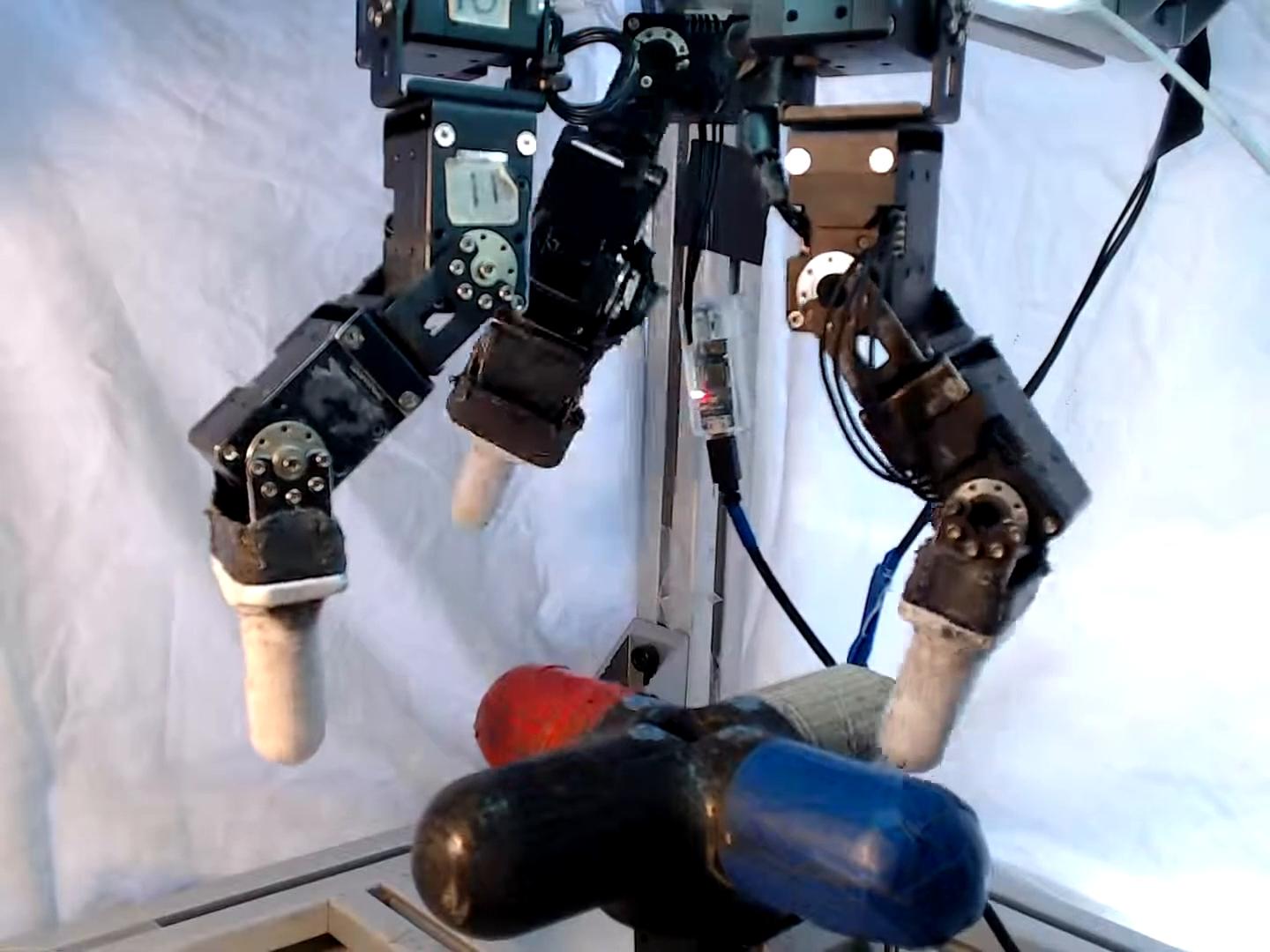}
        \includegraphics[height=0.22\linewidth]{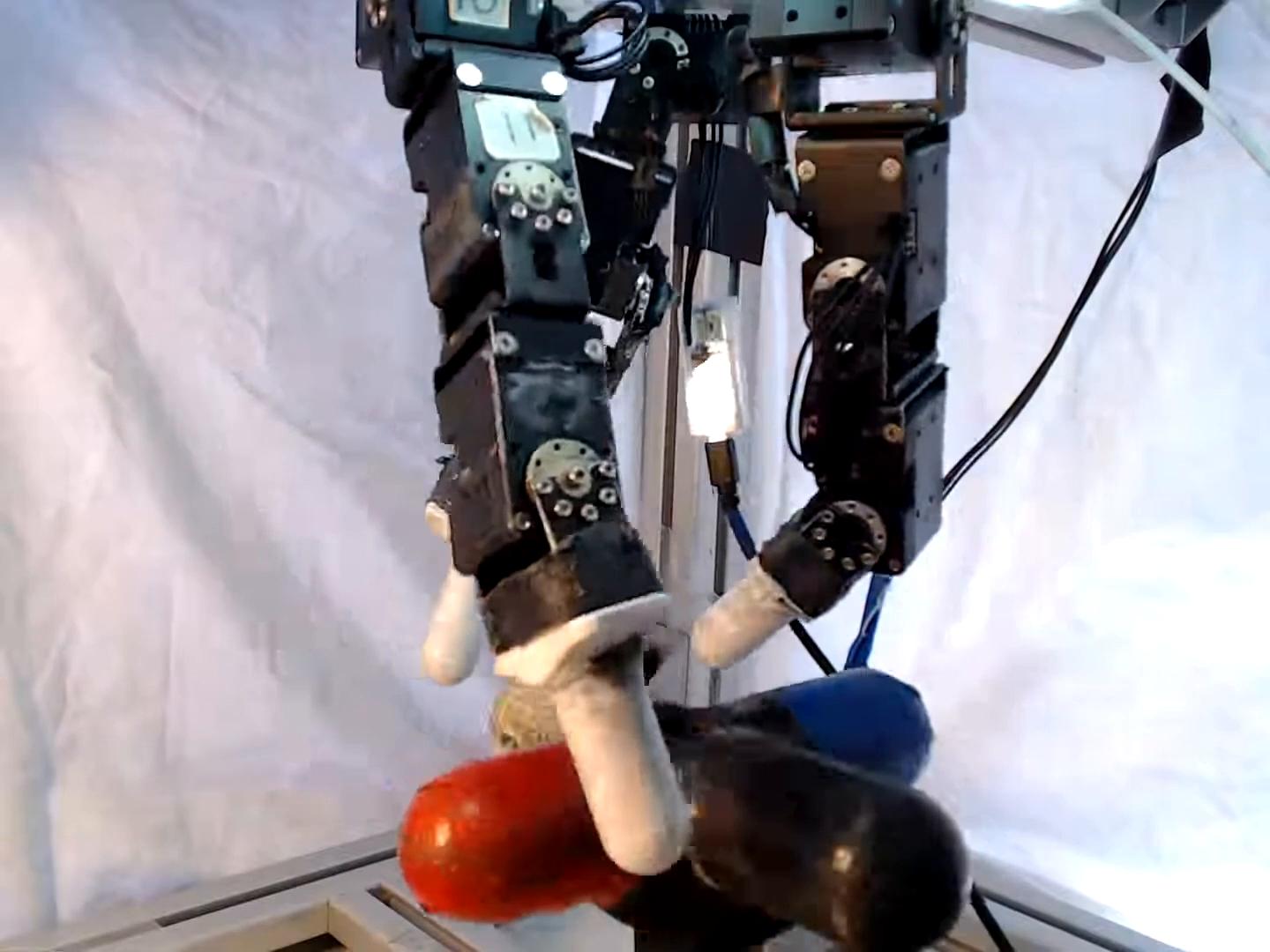}
        \includegraphics[height=0.22\linewidth]{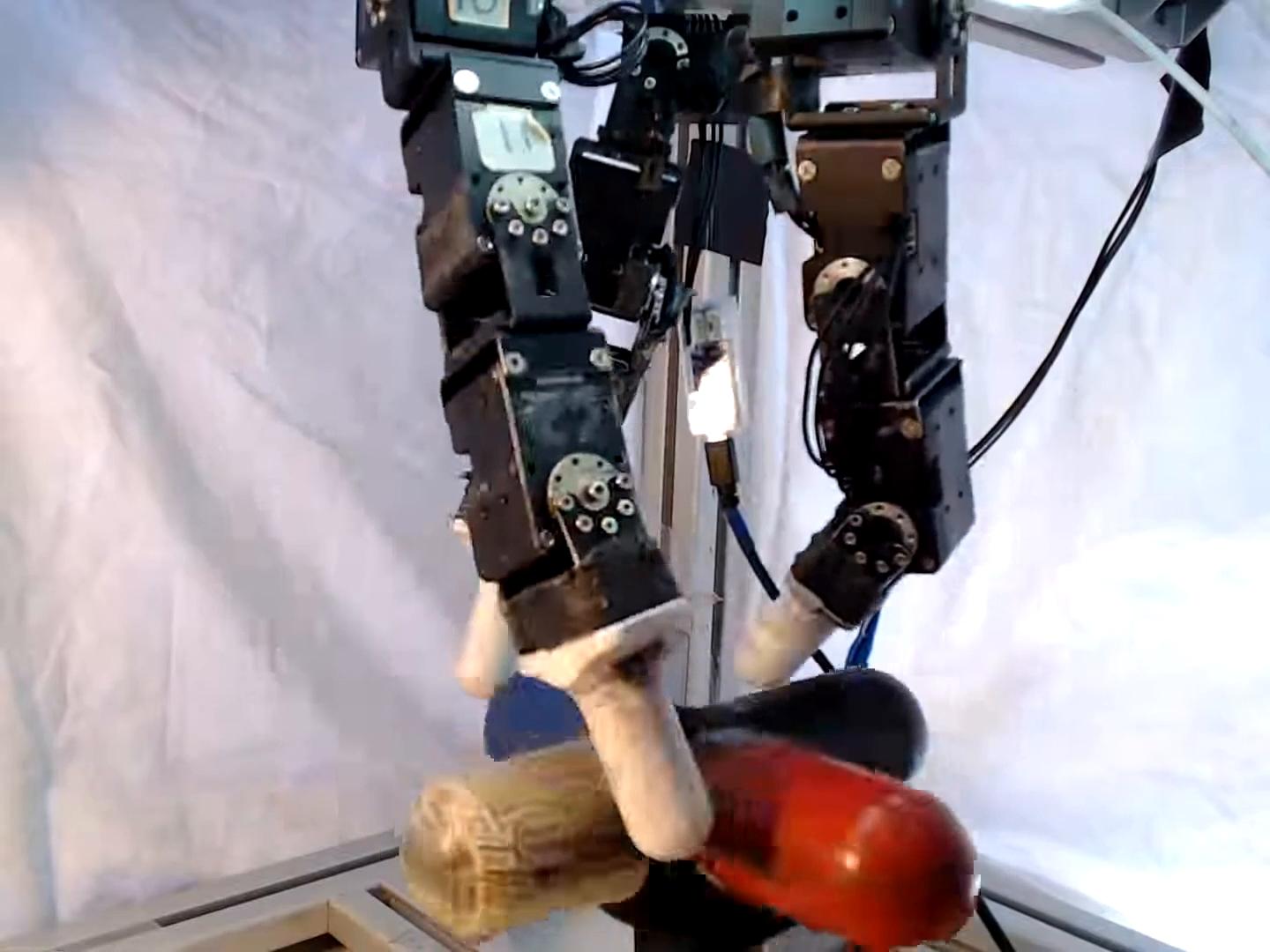}
        \vspace{0.1cm}

        \includegraphics[width=0.99\textwidth]{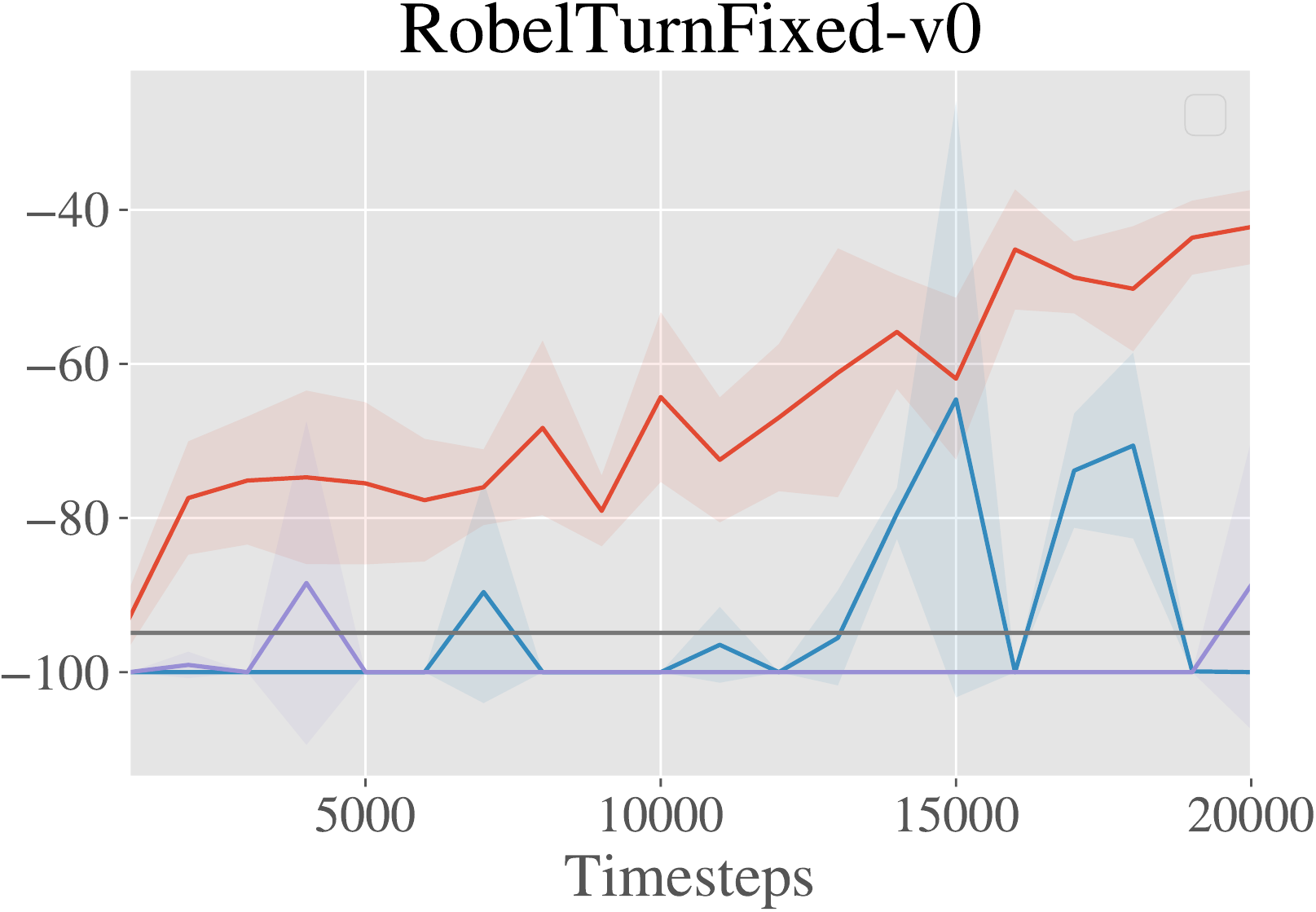}
    \end{subfigure}
    \begin{subfigure}[b]{0.28\textwidth}
        \center
        
        \includegraphics[height=0.22\linewidth]{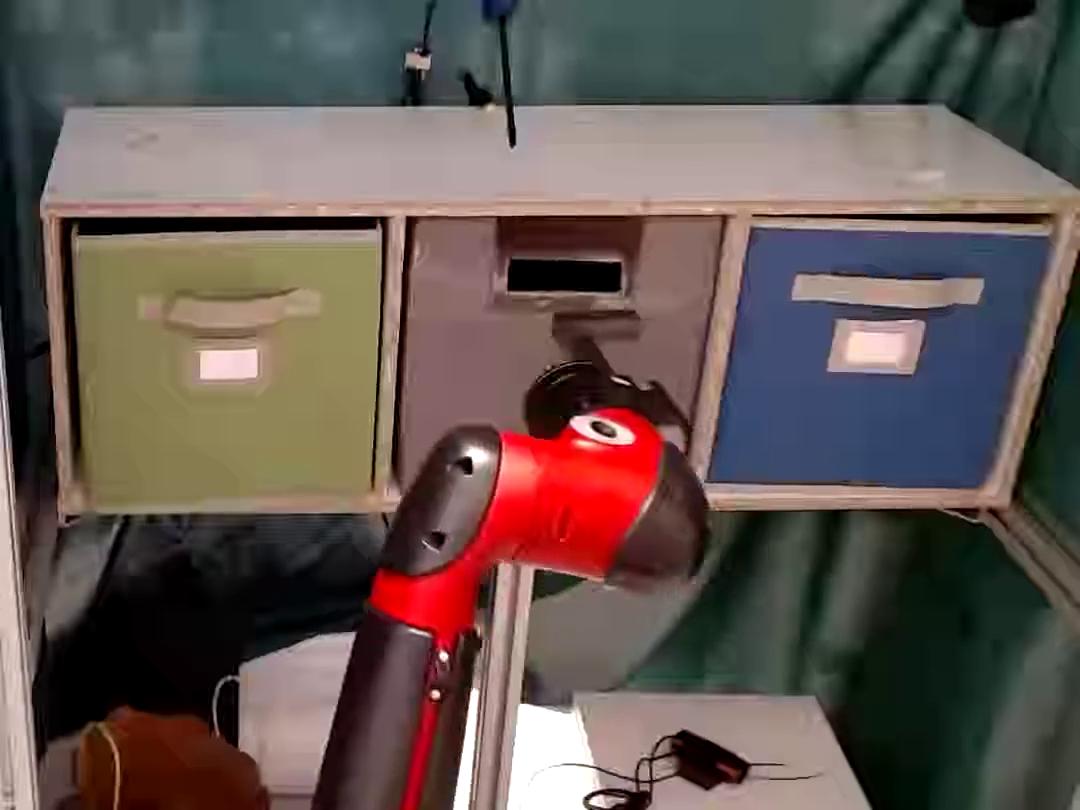}
        \includegraphics[height=0.22\linewidth]{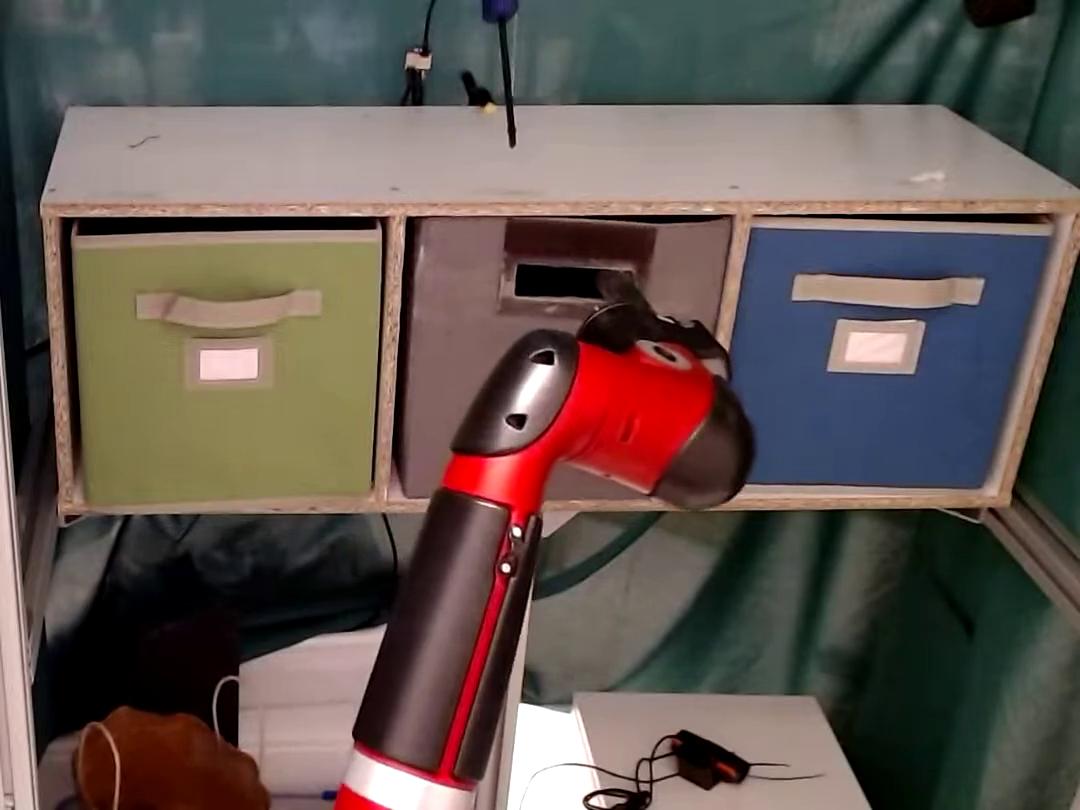}
        \includegraphics[height=0.22\linewidth]{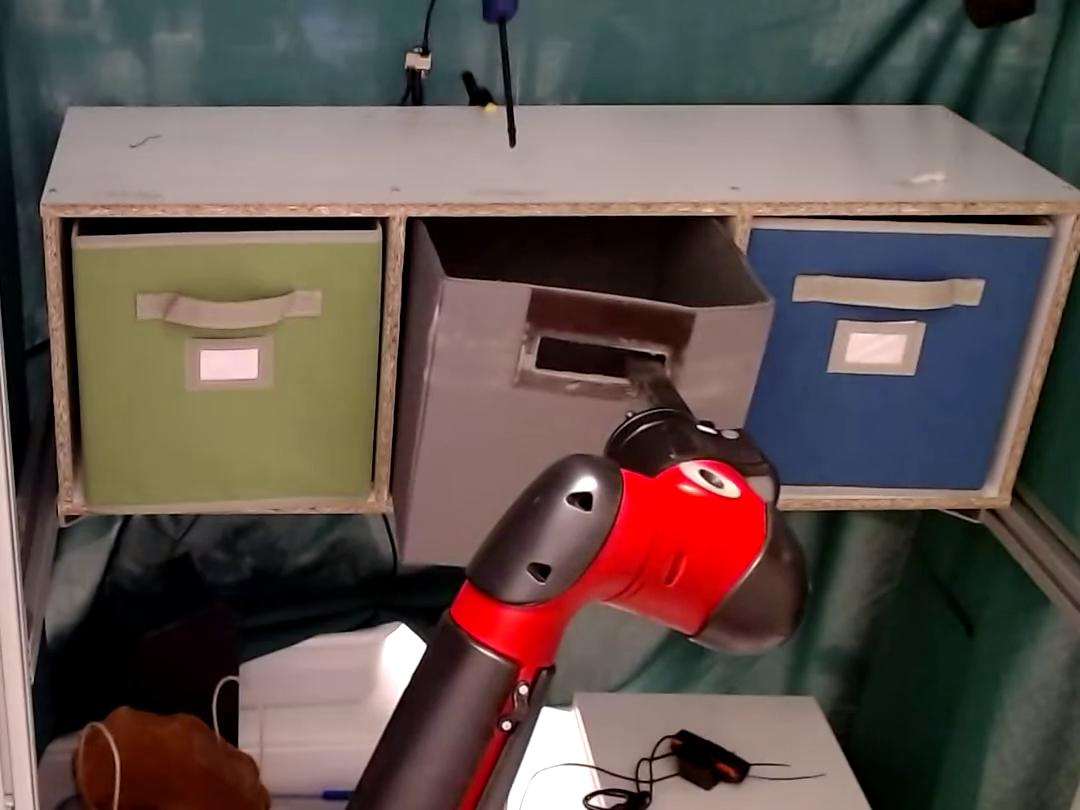}
        \vspace{0.1cm}

        \includegraphics[width=0.99\textwidth]{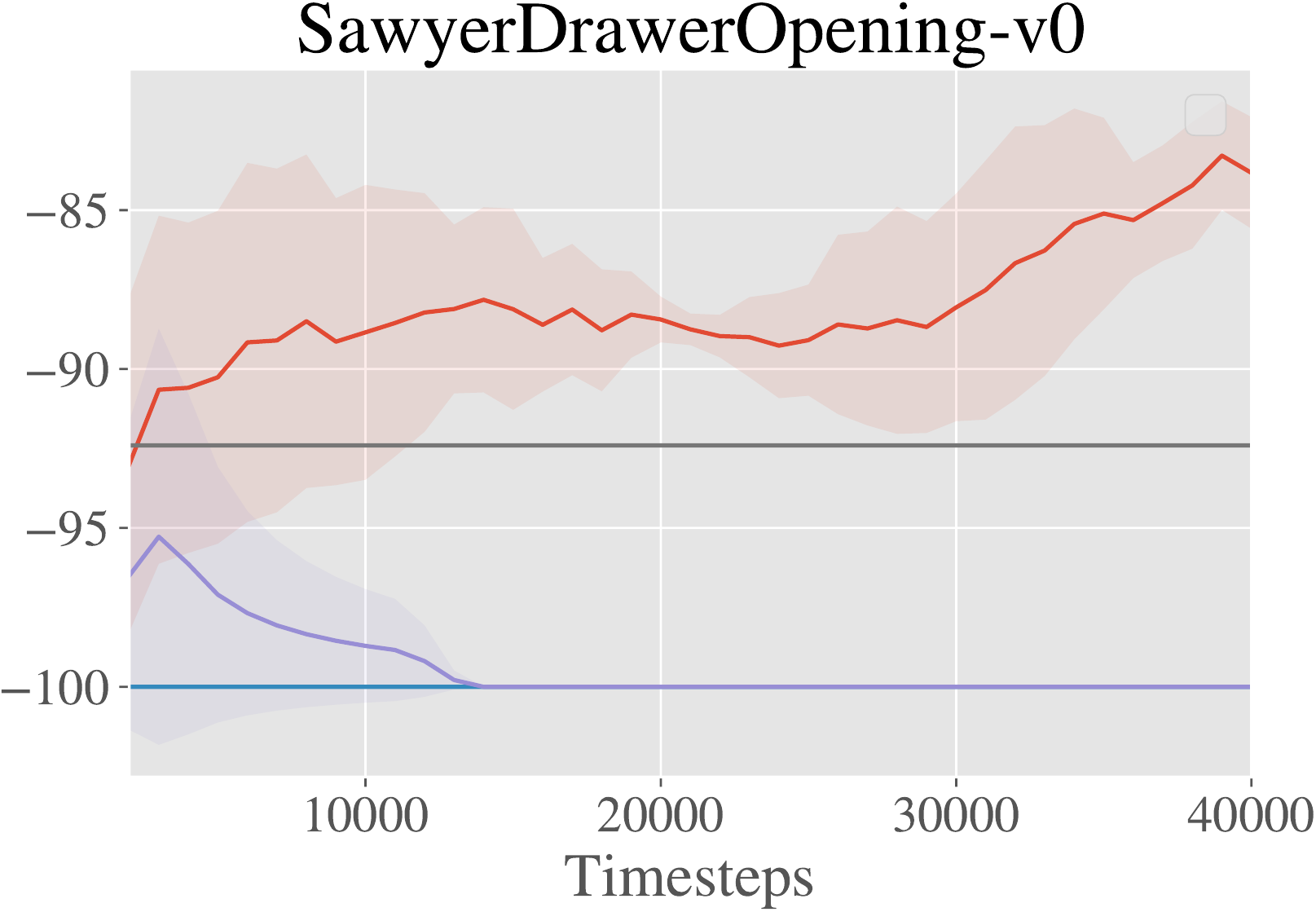}
    \end{subfigure}
    \begin{subfigure}[b]{0.28\textwidth}
        \center
        \hspace{0.02\linewidth}
        \includegraphics[height=0.22\linewidth]{figures/filmstrip_hand/vid_0.jpg}
        \includegraphics[height=0.22\linewidth]{figures/filmstrip_hand/vid_70.jpg}
        \includegraphics[height=0.22\linewidth]{figures/filmstrip_hand/vid_190.jpg}
            \vspace{0.1cm}

        \includegraphics[width=0.99\textwidth]{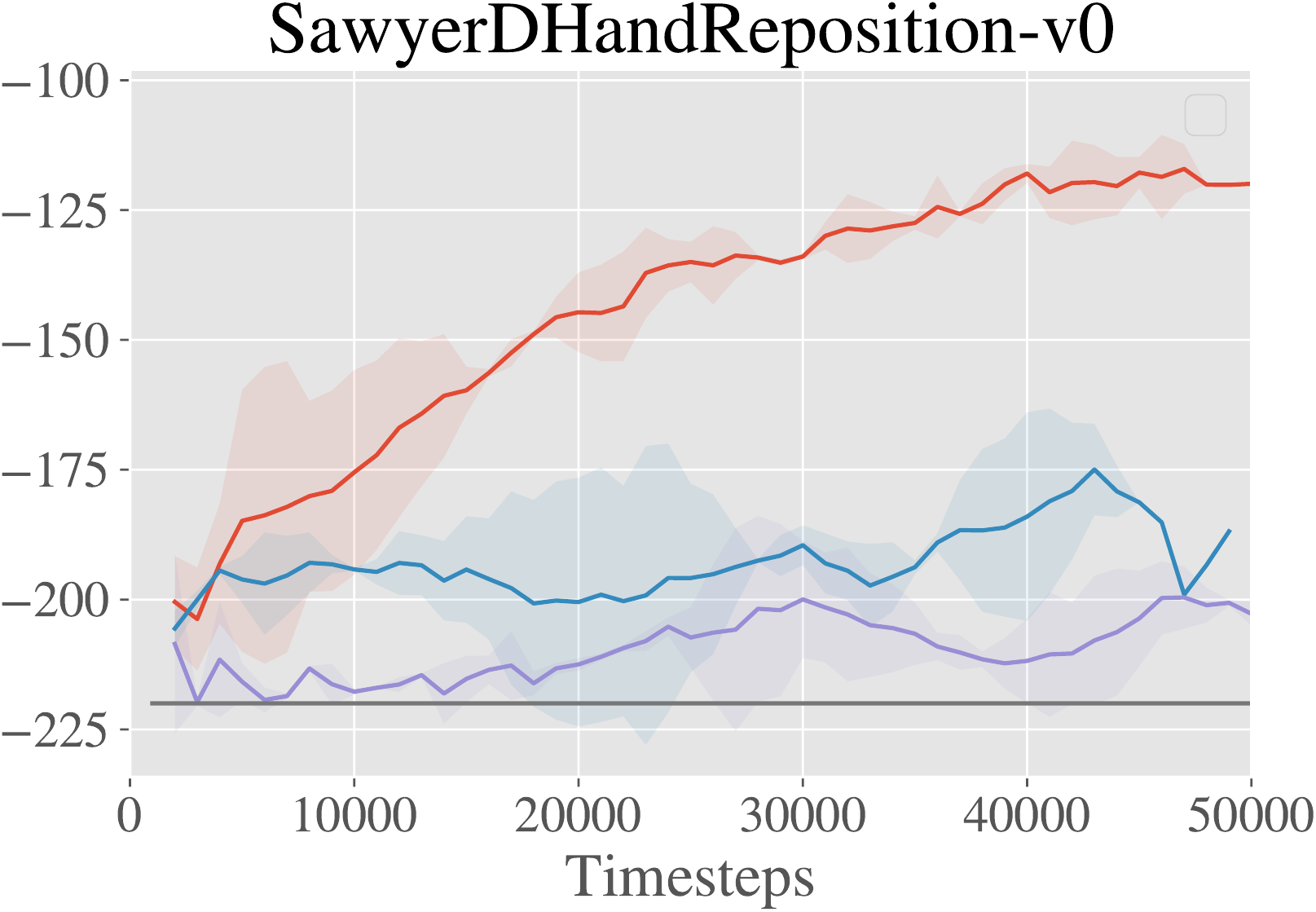}
    \end{subfigure}
    \begin{subfigure}[b]{0.1\textwidth}
        \center
        \hspace{0.02\linewidth}
        \includegraphics[width=0.99\textwidth]{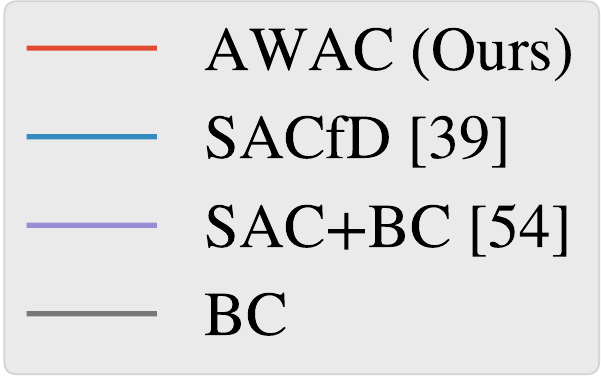}
        \vspace{0.6cm}
    \end{subfigure}

    \caption{
    Algorithm comparison on three real-world robotic environments. Images of real world robotic tasks are pictured above. Left: a three fingered D'claw must rotate a valve $180^\circ$. Middle: a Sawyer robot must slide open a drawer using a hook attachment. Right: a dexterous hand attached to a Sawyer robot must re-position an object to to the center of the table. On each task, AWAC trained offline achieves reasonable performance (shown at timestep 0) and then steadily improves from online interaction. Other methods, which also all have access to prior data, fail to utilize the prior data effectively offline and therefore exhibit slow or no online improvement. Videos of all experiments are available at \projectpage
    \vspace{-0.5cm}
    }
    \label{fig:robot-learning-curves}
\end{figure*}

First, we compare our method on these dexterous manipulation tasks against prior methods for off-policy learning, offline learning, and bootstrapping from demonstrations.
Specific implementation details are discussed in Appendix~\ref{sec:baseline_impl}.
The results are shown in Fig.~\ref{fig:hand-learning-curves}. Our method is able to leverage the prior data to quickly attain good performance, and the efficient off-policy actor-critic component of our approach fine-tunes much more quickly than demonstration augmented policy gradient (DAPG), the method proposed by \citet{rajeswaran2018dextrous}. For example, our method solves the pen task in 120K timesteps, the equivalent of just 20 minutes of online interaction. While the baseline comparisons and ablations make some amount of progress on the pen task, alternative off-policy RL and offline RL algorithms are largely unable to solve the door and relocate task in the time-frame considered. We find that the design decisions to use off-policy critic estimation allow AWAC to outperform AWR~\citep{peng2019awr} while the implicit behavior modeling allows AWAC to significantly outperform ABM~\citep{siegel2020abm}, although ABM does make some progress. \citet{rajeswaran2018dextrous} show that DAPG can solve variants of these tasks with more well-shaped rewards, but requires considerably more samples.

Additionally, we evaluated all methods on the Gym MuJoCo locomotion benchmarks, similarly providing demonstrations as offline data. The results plots for these experiments are included in Appendix~\ref{sec:gym} in the supplementary materials. These tasks are substantially easier than the sparse reward manipulation tasks described above, and a number of prior methods also perform well. 
SAC+BC and BRAC perform on par with our method on the HalfCheetah task, and ABM performs on par with our method on the Ant task, while our method outperforms all others on the Walker2D task. 
However, our method matches or exceeds the best prior method in all cases, whereas no other single prior method attains good performance on all tasks.

\subsubsection{Fine-Tuning from Random Policy Data} \label{sec:random_exps}

\begin{figure}[H]
  \begin{center}
    \begin{subfigure}[b]{0.26\textwidth}
        \includegraphics[width=0.99\textwidth,trim={5cm 0 5cm 0},clip]{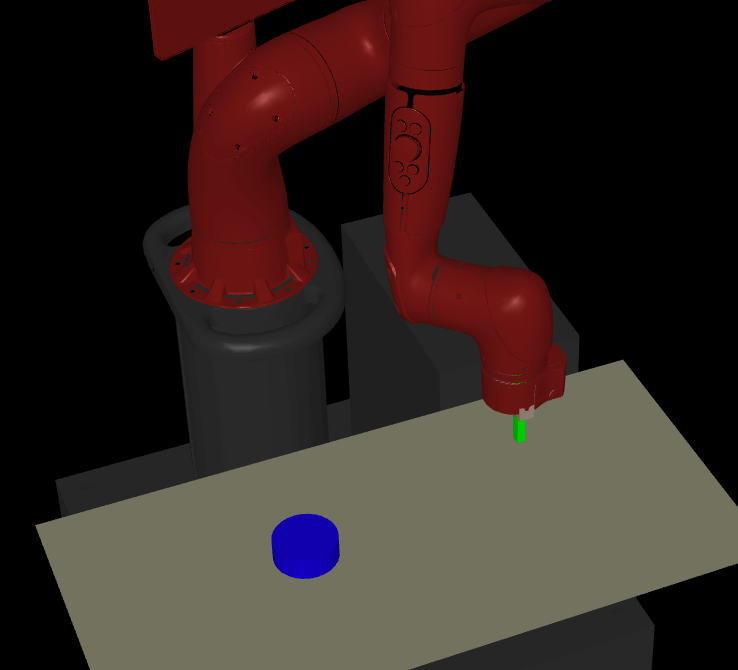}
    \end{subfigure}
    \hspace{0.03\textwidth}
    \begin{subfigure}[b]{0.45\textwidth}
        \includegraphics[width=0.98\textwidth]{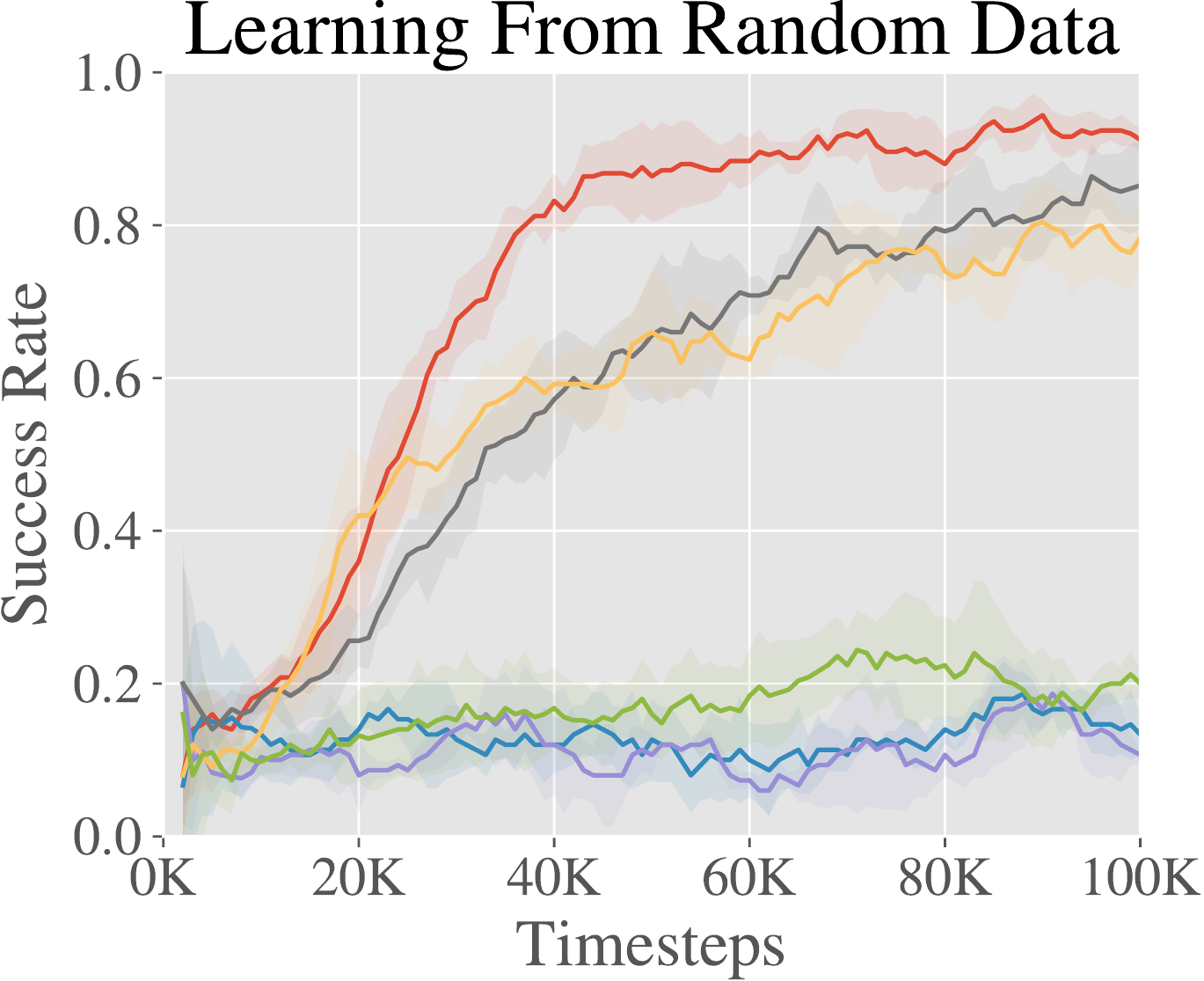}
        
        \includegraphics[width=0.98\textwidth]{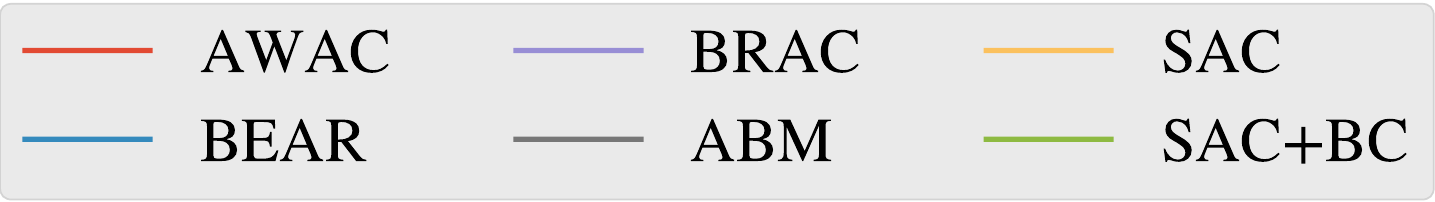}
    \end{subfigure}
  \end{center}
  \vspace{0.1cm}
  \caption{
  Comparison of fine-tuning from an initial dataset of suboptimal data on a Sawyer robot pushing task.
  }
  \vspace{-0.5cm}
 \label{fig:gcrl}
\end{figure}

An advantage of using off-policy RL for reinforcement learning is that we can also incorporate suboptimal data, rather than demonstrations. In this experiment, we evaluate on a simulated tabletop pushing environment with a Sawyer robot pictured in Fig~\ref{fig:hand-learning-curves} and described further in Appendix~\ref{sec:environment_impl}.
To study the potential to learn from suboptimal data, we use an off-policy dataset of 500 trajectories generated by a random process. 
The task is to push an object to a target location in a 40cm x 20cm goal space.
The results are shown in Figure~\ref{fig:gcrl}. We see that while many methods begin at the same initial performance, AWAC learns the fastest online and is actually able to make use of the offline dataset effectively.

\subsection{Real-World Robot Learning with Prior Data} \label{sec:real_world_exps}

We next evaluate AWAC and several baselines on a range of real-world robotic systems, shown in the top row of Fig~\ref{fig:robot-learning-curves}. We study the following tasks:
rotating a valve with a 3-fingered claw, repositioning an object with a 4-fingered hand, and opening a drawer with a Sawyer robotic arm. The dexterous manipulation tasks involve fine finger coordination to properly reorient and reposition objects, as well as high dimensional state and action spaces. The Sawyer
drawer opening task requires accurate arm movements to properly hook the end-effector into the handle of the drawer. To ensure continuous operation, all environments are fitted with an automated reset mechanism that executes before each trajectory is collected, allowing us to run real-world experiments without human supervision.
Since real-world experiments are significantly more time-consuming, we could not compare to the full range of prior methods in the real world, but we include comparisons with the following methods: direct behavioral cloning (BC) of the provided data (which is reasonable in these settings, since the prior data includes demonstrations), off-policy RL with soft actor-critic (SAC)~\cite{haarnoja2018sac}, where the prior data is included in the replay buffer and used to pretrain the policy (which refer to as SACfD), and a modified version of SAC that includes an added behavioral cloning loss (SAC+BC), which is analogous to \citet{nair2018demonstrations} or an off-policy version of \citet{rajeswaran2018dextrous}. Further implementation details of these algorithms are provided in Appendix~\ref{sec:baseline_impl} in the supplementary materials. 

Next, we describe the experimental setup for hardware experiments. Precise details of the hardware setup can be found in Appendix~\ref{sec:hardwaresetup} in the supplementary materials. 

\noindent \textbf{Dexterous Manipulation with a 3-Fingered Claw.}
This task requires controlling a 3-fingered, 9 DoF robotic hand, introduced by~\citet{ahn2019robel}, to rotate a 4-pronged valve object by $180$ degrees. To properly perform this task, multiple fingers need to coordinate to stably and efficiently rotate the valve into the desired orientation. The state space of the system consists of the joint angles of all the 9 joints in the claw, and the action space consists of the joint positions of the fingers,
which are followed by the robot using a low-level PID controller. The reward for this task is sparse: $-1$ if the valve is rotated within 0.25 radians of the target, and $0$ otherwise. Note that this reward function is significantly more difficult than the dense, well-shaped reward function typically used in prior work~\cite{ahn2019robel}. The prior data consists of 10 trajectories collected using kinesthetic teaching, combined this with 200 trajectories obtained through executing a policy trained via imitation learning in the environment.

\noindent \textbf{Drawer Opening with a Sawyer Arm.}
This task requires controlling a Sawyer arm to slide open a drawer. The robot uses 3-dimensional end-effector control, and is equipped with a hook attachment to make the drawer opening possible. The state space is 4-dimensional, consisting of the position of the robot end-effector and the linear position of the drawer, measured using an encoder. The reward is sparse: $-1$ if the drawer is open beyond a threshold and $0$ otherwise. For this task, the prior data consists of 10 demonstration trajectories collected using via teleoperation with a 3D mouse, as well as 500 trajectories obtained through executing a policy trained via imitation learning in the environment. This task is challenging because it requires very precise insertion of the hook attachment into the opening, as pictured in Fig~\ref{fig:robot-learning-curves}, before the robot can open the drawer. Due to the sparse reward, making learning progress on this task requires utilizing prior data to construct an initial policy that at least sometimes succeeds.

\noindent \textbf{Dexterous Manipulation with a Robotic Hand.}
This task requires controlling a 4-fingered robotic hand mounted on a Sawyer robotic arm to reposition an object~\cite{gupta2021dexterous}. The task requires careful coordination between the hand and the arm to manipulate the object accurately. The reward for this task is a combination of the negative distance between the hand and the object and the negative distance between the object and the target. The actions are 19-dimensional, consisting of 16-dimensional finger control and 3-dimensional end effector control of the arm. For this task, the prior data of $19$ trajectories were collected using kinesthetic teaching and combined with $50$ trajectories obtained by executing a policy trained with imitation learning on this data.

The results on these tasks are shown in Figure \ref{fig:robot-learning-curves}. We first see that AWAC attains performance that is comparable to the best prior method from offline training alone, as indicated by the value at time step 0 (which corresponds to the beginning of online finetuning). This means that, during online interaction, AWAC collects data that is of higher quality, and improves more quickly. The prior methods struggle to improve from online training on these tasks, likely because the sparse reward function and challenging dynamics make progress very difficult from a bad initialization. These results suggest that AWAC is effectively able to leverage prior data to bootstrap online reinforcement learning in the real world, even on tasks with difficult and uninformative reward functions.

\section{Discussion and Future Work}\label{sec:conclusion}

We have discussed the challenges existing RL methods face when fine-tuning from prior datasets, and proposed an algorithm, AWAC, for this setting. The key insight in AWAC is that an implicitly constrained actor-critic algorithm is able to both train offline and continue to improve with more experience. We provide detailed empirical analysis of the design decisions behind AWAC, showing the importance of off-policy learning, bootstrapping and the particular form of implicit constraint enforcement. To validate these ideas, we evaluate on a variety of simulated and real world robotic problems. 

While AWAC is able to effectively leverage prior data for significantly accelerating learning, it does run into some limitations. Firstly, it can be challenging to choose the appropriate threshold for constrained optimization. Resolving this would involve exploring adaptive threshold tuning schemes. Secondly, while AWAC is able to avoid over-conservative behavior empirically, in future work, we hope to analyze theoretical factors that go into building a good finetuning algorithm. And lastly, in the future we plan on applying AWAC to more broadly incorporate data across different robots, labs and tasks rather than just on isolated setups. By doing so, we hope to enable an even wider array of robotic applications.

\section{Acknowledgements}
This research was supported by the Office of Naval Research, the National Science Foundation through IIS-1700696 and IIS-1651843, and ARL DCIST CRA W911NF-17-2-0181. We would like to thank Aviral Kumar, Ignasi Clavera, Karol Hausman, Oleh Rybkin, Michael Chang, Corey Lynch, Kamyar Ghasemipour, Alex Irpan, Vitchyr Pong, Graham Hughes, Zihao Zhao, Vikash Kumar, Saurabh Gupta, Shubham Tulsiani, Abhinav Gupta and many others at UC Berkeley RAIL Lab and Robotics at Google for their valuable feedback on the paper and insightful discussions.

\raggedbottom

\bibliography{main}
\bibliographystyle{iclr2021_conference}

\clearpage
\newpage

\appendix

\section{Appendix}

\subsection{Algorithm Derivation Details} \label{sec:derivation}

The full optimization problem we solve, given the previous off-policy advantage estimate $A^{\pi_k}$ and buffer distribution $\pi_\beta$, is given below:
\begin{align}
    \pinew = \; \argmax_{\pi \in \Pi} \; & \E_{\at \sim \pi(\cdot|\st)}[A^{\pi_k}(\st, \at)] \\
    \text{s.t.} \; & \KL(\pi(\cdot|\st)||\pi_\buffer(\cdot|\st)) \leq \epsilon \\
    & \int_\at \pi(\at|\st) d\at = 1.
\end{align}
Our derivation follows~\citet{peters2010reps} and~\citet{peng2019awr}. The analytic solution for the constrained optimization problem above can be obtained by enforcing the KKT conditions. The Lagrangian is:
\begin{align}
    \mathcal{L}(\pi, \lagrangeawr, \alpha) = & \E_{\at \sim \pi(\cdot|\st)}[A^{\pi_k}(\st, \at)] \\ & + \lagrangeawr(\epsilon - \KL(\pi(\cdot|\st)||\piold(\cdot|\st))) \\ & + \alpha (1 - \int_\at \pi(\at|\st) d\at).
\end{align}
Differentiating with respect to $\pi$ gives:
\begin{align}
    \frac{\partial \mathcal{L}}{\partial \pi} = A^{\pi_k}(\st, \at) - \lagrangeawr \log \pi_\beta(\at|\st) + \lagrangeawr \log \pi(\at|\st) + \lagrangeawr - \alpha.
\end{align}
Setting $\frac{\partial \mathcal{L}}{\partial \pi}$ to zero and solving for $\pi$ gives the closed form solution to this problem:
\begin{align}
    \label{eqn:closedform}
    \pi^*(\at|\st) = \frac{1}{Z(\st)} \pi_\buffer(\at|\st)\exp\left(\frac{1}{\lagrangeawr} A^{\pi_k}(\st, \at)\right),
\end{align}
Next, we project the solution into the space of parametric policies. For a policy $\pi_\theta$ with parameters $\theta$, this can be done by minimizing the KL divergence of $\pi_{\theta}$ from the optimal non-parametric solution $\pi^*$ under the data distribution $\rho_{\pi_\buffer}(\st)$: 
\begin{align}
    & \argmin_\theta \; \Ex_{\rho_{\pi_\buffer}(\st) } \left[\KL(\pi^*(\cdot|\st)||\pi_\theta(\cdot|\st))\right] \\ = & 
    \argmin_\theta \; \Ex_{\rho_{\pi_\buffer}(\st)} \left[\Ex_{\pi^*(\cdot|\st)}[-\log \pi_\theta(\cdot|\st)]\right]
\end{align}
Note that in the projection step, the parametric policy could be projected with either direction of KL divergence. However, choosing the reverse KL direction has a key advantage: it allows us to optimize $\theta$ as a maximum likelihood problem with an expectation over data $s, a \sim \buffer$, rather than sampling actions from the policy that may be out of distribution for the Q function. In our experiments we show that this decision is vital for stable off-policy learning.

Furthermore, assume discrete policies with a minimum probably density of $\pi_\theta \geq \alpha_\theta$. Then the upper bound:
\begin{align}
    \KL(\pi^*||\pi_\theta) \leq & \frac{2}{\alpha_\theta} \DTV(\pi^*, \pi_\theta)^2 \\
    \leq & \frac{1}{\alpha_\theta} \KL(\pi_\theta||\pi^*)
\end{align}
holds by the Pinsker's inequality, where $\DTV$ denotes the total variation distance between distributions. Thus minimizing the reverse KL also bounds the forward KL. Note that we can control the minimum $\alpha$ if desired by applying Laplace smoothing to the policy.

\subsection{Implementation Details} \label{sec:implementation}

We implement the algorithm building on top of twin soft actor-critic~\citep{haarnoja2018sac}, which incorporates the twin Q-function architecture from twin delayed deep deterministic policy gradient (TD3) from~\citet{fujimoto2018td3}. All off-policy algorithm comparisons (SAC, BRAC, MPO, ABM, BEAR) are implemented from the same skeleton. The base hyperparameters are given in Table~\ref{table:rl-hyperparams}. The policy update is replaced with:
\begin{align}
    \theta_{k+1} = \argmax_\theta \; & \; \Ex_{\st, \at \sim \buffer}
    \left[\log \pi_\theta(\at|\st) \frac{1}{Z(\st)}  \exp \left(\frac{1}{\lagrangeawr}A^{\pi_k}(\st, \at) \right)\right].
\end{align}

\begin{table}[h!]
\footnotesize
\begin{tabular}{c|c|c}
Env      & \shortstack{Use \\ $Z(\st)$} & \shortstack{Omit \\ $Z(\st)$} \\ \hline
pen      & 84\%      & 98\%    \\
door     & 0\%      & 95\%    \\
relocate & 0\%      & 54\%
\end{tabular}
\caption{Success rates after online fine-tuning (after 800K steps for pen, door and 4M steps for relocate) using AWAC with and without $Z(\st)$ weight. These results show that although we can estimate $Z(\st)$, weighting by $Z(\st)$ actually results in worse performance.}
\label{fig:z}
\end{table}

Similar to advantage weight regression~\citep{peng2019awr} and other prior work~\citep{neumann2008fqiawr, wang2018marwil, siegel2020abm},
we disregard the per-state normalizing constant $Z(\st) = \int_\at \pi_\theta(\at|\st) \exp \left(\frac{1}{\lagrangeawr}A^{\pi_k}(\st, \at)\right) d\at = \E_{\at \sim \pi_\theta(\cdot|\st)}[A^{\pi_k}(\st, \at)]$. We did experiment with estimating this expectation per batch element with $K=10$ samples,
but found that this generally made performance worse, perhaps because errors in the estimation of $Z(\st)$ caused more harm than the benefit the method derived from estimating this value. We report success rate results for variants of our method with and without $Z(\st)$ estimation in Table~\ref{fig:z}.

While prior work~\citep{neumann2008fqiawr, wang2018marwil, peng2019awr} has generally ignored the omission of $Z(\st)$ without any specific justification, it is possible to bound this value both above and below using the Cauchy-Schwarz and reverse Cauchy-Schwarz (Polya-Szego) inequalities, as follows.
Let $f(\at) = \pi(\at|\st)$ and $g(\at) = \exp(A(\st, \at)/\lambda)$. Note $f(\at) > 0$ for stochastic policies and $g(\at) > 0$.
By Cauchy-Schwarz, $Z(s) = \int_\at f(\at) g(\at) d\at \leq \sqrt{\int_\at f(\at)^2 d\at \int_\at g(\at)^2 d\at} = C_1$. To apply Polya-Szego, let $m_f$ and $m_g$ be the minimum of $f$ and $g$ respectively and $M_f, M_g$ be the maximum. Then $Z(\st) \geq 2 (\sqrt{\frac{M_f M_g}{m_f m_g} + \frac{m_f m_g}{M_f M_g}})^{-1} C_1 = C_2$. We therefore have $C_1 \leq Z(\st) \leq C_2$, though the bounds are generally not tight.

A further, more intuitive argument for why omitting $Z(\st)$ may be harmless in practice comes from observing that this normalizing factor only affects the relative weight of different \emph{states} in the training objective, not different actions. The state distribution in $\beta$ already differs from the distribution over states that will be visited by $\pi_\theta$, and therefore preserving this state distribution is likely to be of limited utility to downstream policy performance. Indeed, we would expect that sufficiently expressive policies would be less affected by small to moderate variability in the state weights. On the other hand, inaccurate estimates of $Z(\st)$ may throw off the training objective by increasing variance, similar to the effect of degenerate importance weights.


The Lagrange multiplier $\lagrangeawr$ is treated as a hyperparameter in our method. In this work we use $\lagrangeawr = 0.3$ for the manipulation environments and $\lagrangeawr = 1.0$ for the MuJoCo benchmark environments. One could adaptively learn $\lagrangeawr$ with a dual gradient descent procedure, but this would require access to $\pi_\beta$.

As rewards for the dextrous manipulation environments are non-positive, we clamp the Q value for these experiments to be at most zero. We find this stabilizes training slightly.

\begin{table}
    \centering
    \begin{tabular}{c|c}
    \hline
    \textbf{Hyper-parameter} & \textbf{Value} \\
    \hline
    Training Batches Per Timestep & $1$\\
    Exploration Noise & None (stochastic policy) \\
    RL Batch Size & $1024$ \\
    Discount Factor & $0.99$\\
    Reward Scaling & $1$\\
    Replay Buffer Size & $1000000$\\
    Number of pretraining steps & $25000$ \\
    Policy Hidden Sizes & $[256, 256, 256, 256]$\\
    Policy Hidden Activation & ReLU\\
    Policy Weight Decay & $10^{-4}$ \\
    Policy Learning Rate & $3 \times 10^{-4}$\\
    Q Hidden Sizes & $[256, 256, 256, 256]$\\
    Q Hidden Activation & ReLU\\
    Q Weight Decay & $0$ \\
    Q Learning Rate & $3 \times 10^{-4}$\\
    Target Network $\tau$ & $5\times10^{-3}$ \\
    \hline
    \end{tabular}
\caption{Hyper-parameters used for RL experiments.}
\label{table:rl-hyperparams}
\end{table}

\subsection{Environment-Specific Details} \label{sec:environment_impl}

We evaluate our method on three domains: dexterous manipulation environments, Sawyer manipulation environments, and MuJoCo benchmark environments. In the following sections we describe specific details.

\subsubsection{Dexterous Manipulation Environments}

These environments are modified from those proposed by~\citet{rajeswaran2018dextrous}.

\paragraph{pen-binary-v0.} The task is to spin a pen into a given orientation. The action dimension is 24 and the observation dimension is 45. Let the position and orientation of the pen be denoted by $x_p$ and $x_o$ respectively, and the desired position and orientation be denoted by $d_p$ and $d_o$ respectively. The reward function is $r = \mathds{1}_{|x_p - d_p| \leq 0.075} \mathds{1}_{|x_o \cdot d_o| \leq 0.95}$ - 1.
In~\citet{rajeswaran2018dextrous}, the episode was terminated when the pen fell out of the hand; we did not include this early termination condition.

\paragraph{door-binary-v0.} The task is to open a door, which requires first twisting a latch. The action dimension is 28 and the observation dimension is 39. Let $d$ denote the angle of the door. The reward function is $r = \mathds{1}_{d > 1.4}$ - 1.

\paragraph{relocate-binary-v0.} The task is to relocate an object to a goal location. The action dimension is 30 and the observation dimension is 39. Let $x_p$ denote the object position and $d_p$ denote the desired position. The reward is $r = \mathds{1}_{|x_p - d_p| \leq 0.1}$ - 1.

\subsubsection{Sawyer Manipulation Environment}

\paragraph{SawyerPush-v0.} This environment is included in the \href{https://github.com/vitchyr/multiworld}{Multiworld} library. The task is to push a puck to a goal position in a 40cm x 20cm, and the reward function is the negative distance between the puck and goal position. When using this environment, we use hindsight experience replay for goal-conditioned reinforcement learning. The random dataset for prior data was collected by rolling out an Ornstein-Uhlenbeck process with $\theta = 0.15$ and $\sigma = 0.3$.

\subsubsection{Off-Policy Data Performance}

The performances of the expert data, behavior cloning (BC) on the expert data (1), and BC on the combined expert+BC data (2) are included in Table~\ref{fig:bc}. For Gym benchmarks we report average return, and expert data is collected by a trained SAC policy. For dextrous manipulation tasks we report the success rate, and the expert data consists of human demonstrations provided by~\citet{rajeswaran2018dextrous}.

\begin{table}[h!]
\footnotesize
\begin{tabular}{c|c|c|c}
Env         & Expert & BC (1) & BC (2) \\ \hline
cheetah & 9962   & 2507         & 4524      \\
walker      & 5062   & 2040         & 1701      \\
ant         & 5207   & 687          & 1704      \\
pen         & 1      & 0.73          & 0.76       \\
door        & 1      & 0.10          & 0.00       \\
relocate    & 1      & 0.02          & 0.01       \\
\end{tabular}
\caption{Performance of the off-policy data for each environment. BC (1) indicates BC on the expert data, while BC (2) indicates BC on the combined expert+BC data used as off-policy data for pretraining. }
\label{fig:bc}
\end{table}

\begin{figure}[H]
    \begin{centering}
        \footnotesize
        \begin{tabular}{ l||l|l|l|l  }
            Name & $\hat{Q}$ & Policy Objective & $\hat{\pi}_\beta$? & Constraint \\
            \hline
            SAC    & $Q^\pi$   & $\KL(\pi_\theta||\bar{Q})$   & No   &  None            \\
            SAC + BC & $Q^\pi$   & Mixed   & No   &  None            \\
            BCQ    & $Q^\pi$   & $\KL(\pi_\theta||\bar{Q})$   & Yes   &  Support ($\ell^\infty$)  \\
            BEAR   & $Q^\pi$   & $\KL(\pi_\theta||\bar{Q})$   & Yes   &  Support (MMD)   \\
            AWR    & $Q^\beta$ & $\KL(\bar{Q}||\pi_\theta)$   & No   &  Implicit        \\
            MPO    & $Q^\pi$   & $\KL(\bar{Q}||\pi_\theta)$   & Yes$^*$   &  Prior   \\
            ABM-MPO    & $Q^\pi$   & $\KL(\bar{Q}||\pi_\theta)$   & Yes   &  Learned Prior   \\
            DAPG   & -         & $J(\pi_\theta)$   & No   &  None            \\
            BRAC   & $Q^\pi$   & $\KL(\pi_\theta||\bar{Q})$   & Yes   &  Explicit KL penalty   \\
            AWAC (Ours)   & $Q^\pi$   & $\KL(\bar{Q}||\pi_\theta)$  & No    &  Implicit
        \end{tabular}
    \end{centering}
    
    \caption{Comparison of prior algorithms that can incorporate prior datasets. See section~\ref{sec:baseline_impl} for specific implementation details. We argue that avoiding estimating $\hat{\pi}_\beta$ (i.e., $\hat{\pi}_\beta$ is ``No'') is important when learning with complex datasets that include experience from multiple policies, as in the case of online fine-tuning, and maintaining a constraint of some sort is essential for offline training. At the same time, sample-efficient learning requires using $Q^\pi$ for the critic. Our algorithm is the only one that fulfills all of these requirements.}
    \label{fig:algo_table}
\end{figure}

\subsection{Baseline Implementation Details} \label{sec:baseline_impl}

We used public implementations of prior methods (DAPG, AWR) when available. We implemented the remaining algorithms in our framework, which also allows us to understand the effects of changing individual components of the method. In the section, we describe the implementation details. The full overview of algorithms is given in Figure~\ref{fig:algo_table}.

\textbf{Behavior Cloning (BC).}  This method learns a policy with supervised learning on demonstration data.

\textbf{Soft Actor Critic (SAC).} Using the soft actor critic algorithm from \citep{haarnoja2018sac}, we follow the exact same procedure as our method in order to incorporate prior data, initializing the policy with behavior cloning on demonstrations and adding all prior data to the replay buffer. 

\textbf{Behavior Regularized Actor Critic (BRAC).} We implement BRAC as described in \citep{wu2019brac} by adding policy regularization $\log(\pi_\beta(a|s))$ where $\pi_\beta$ is a behavior policy trained with supervised learning on the replay buffer. We add all prior data to the replay buffer before online training. 

\textbf{Advantage Weighted Regression (AWR).} Using the advantage weighted regression algorithm from \citep{peng2019awr}, we add all prior data to the replay buffer before online training. We use the implementation provided by \citet{peng2019awr}, with the key difference from our method being that AWR uses TD($\lambda$) on the replay buffer for policy evaluation.

\textbf{Monotonic Advantage Re-Weighted Imitation Learning (MARWIL).} Monotonic advantage re-weighted imitation learning was proposed by~\citet{wang2018marwil} for offline imitation learning. MARWIL was not demonstrated in online RL settings, but we evaluate it for offline pretraining followed by online fine-tuning as we do other offline algorithms. Although derived differently, MARWIL and AWR are similar algorithms and only differ in value estimation: MARWIL uses the on-policy single-path advantage estimate $A(s, a) = Q^{\pi_\beta}(s, a) - V^{\pi_\beta}(s)$ instead of TD($\lambda$) as in AWR. Thus, we implement MARWIL by modifying the implementation of AWR.

\textbf{Maximum a Posteriori Policy Optimization (MPO).} We evaluate the MPO algorithm presented by~\citet{we2018mpo}. Due to a public implementation being unavailable, we modify our algorithm to be as close to MPO as possible. In particular, we change the policy update in \METHOD to be: 
\begin{align}
    \theta_i & \longleftarrow \argmax_{\theta_i} \; \; \E_{s \sim \mathcal{D}, a \sim \pi(a|s)} \; \nonumber \\ 
    &\left[\log \pi_{\theta_i}(a|s) \exp(\frac{1}{\beta}Q^\piold(s, a))\right].
\end{align}
Note that in MPO, actions for the update are sampled from the policy and the Q-function is used instead of advantage for weights. We failed to see offline or online improvement with this implementation in most environments, so we omit this comparison in favor of ABM.

\textbf{Advantage-Weighted Behavior Model (ABM).} We evaluate ABM, the method developed in \citet{siegel2020abm}. As with MPO, we modify our method to implement ABM, as there is no public implementation of the method. ABM first trains an advantage model $\pi_{\theta_\text{abm}}(a|s)$:
\begin{align}
    \theta_\text{abm} &= \argmax_{\theta_i} \; \; \E_{\tau \sim \mathcal{D}} \nonumber \\ &\left[\sum_{t=1}^{|\tau|} \log \pi_{\theta_\text{abm}}(a_t|s_t)f(R(\tau_{t:N})-\hat{V}(s))\right].
\end{align}
where $f$ is an increasing non-negative function, chosen to be $f = 1_+$. In place of an advantage computed by empirical returns $R(\tau_{t:N})-\hat{V}(s)$ we use the advantage estimate computed per transition by the $Q$ value $Q(s, a)-V(s)$. This is favorable for running ABM online, as computing $R(\tau_{t:N})-\hat{V}(s)$ is similar to AWR, which shows slow online improvement. We then use the policy update: 
\begin{align}
    \theta_i & \longleftarrow \argmax_{\theta_i} \; \; \E_{s \sim \mathcal{D}, a \sim \pi_\text{abm}(a|s)} \nonumber \\
    &\left[\log \pi_{\theta_i}(a|s)  \exp\left(\frac{1}{\lagrangeawr}(Q^{\pi_i}(s, a)-V^{\pi_i}(s))\right)\right].
\end{align}
Additionally, for this method, actions for the update are sampled from a behavior policy trained to match the replay buffer and the value function is computed as $V^\pi(s) = Q^\pi(s, a)$ s.t. $a \sim \pi$.

\textbf{Demonstration Augmented Policy Gradient (DAPG).} We directly utilize the code provided in ~\citep{rajeswaran2018dextrous} to compare against our method. Since DAPG is an on-policy method, we only provide the demonstration data to the DAPG code to bootstrap the initial policy from.

\begin{figure*}[t]
    \centering
    \begin{subfigure}[b]{0.02\textwidth}
        \center
        \begin{turn}{90} 
            \footnotesize
            Average Return
        \end{turn}
        \vspace{1.5cm}
    \end{subfigure}
    \begin{subfigure}[b]{0.3\textwidth}
        \center
        \includegraphics[width=\textwidth]{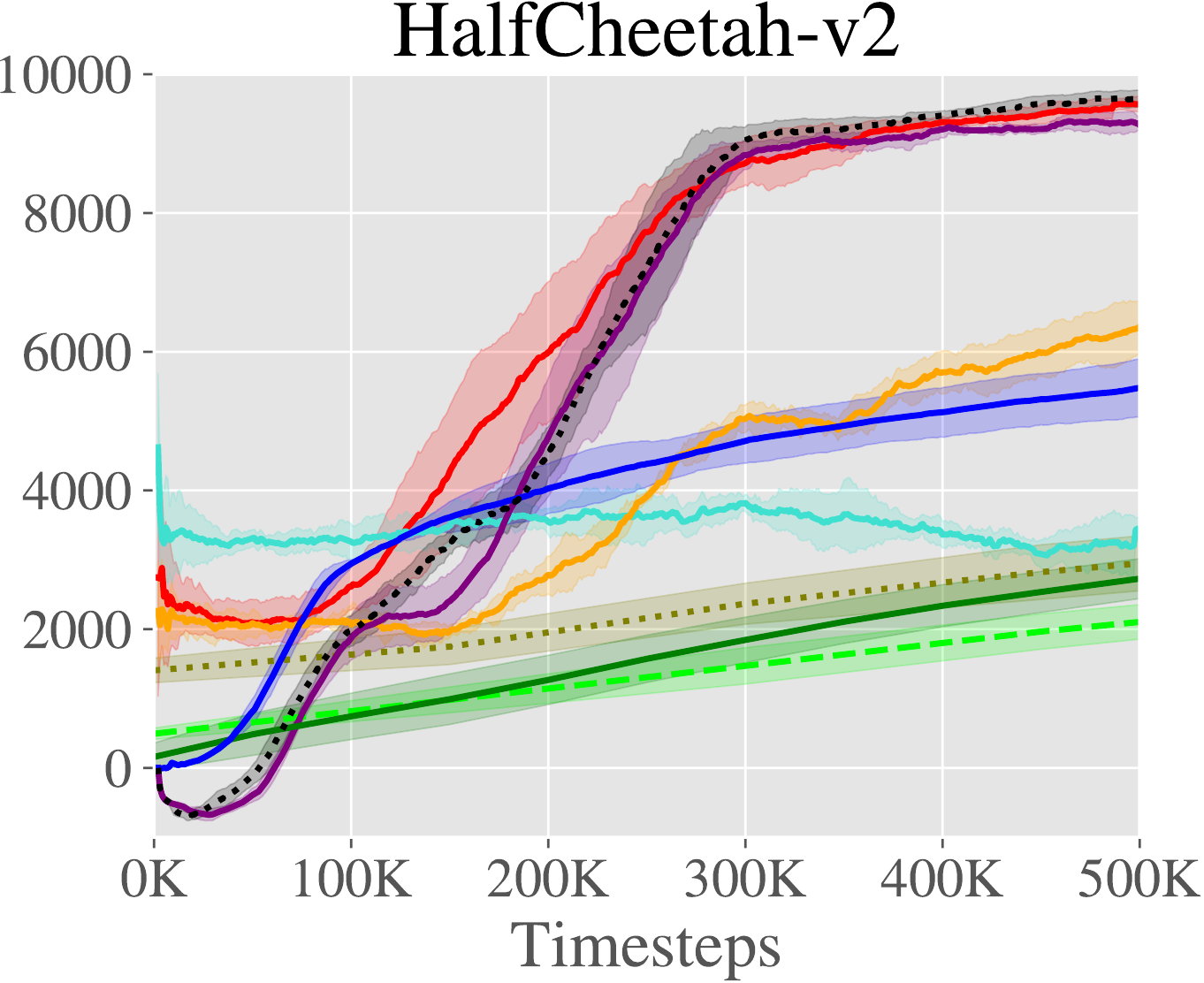}
    \end{subfigure}
    \begin{subfigure}[b]{0.3\textwidth}
        \center
        \includegraphics[width=\textwidth]{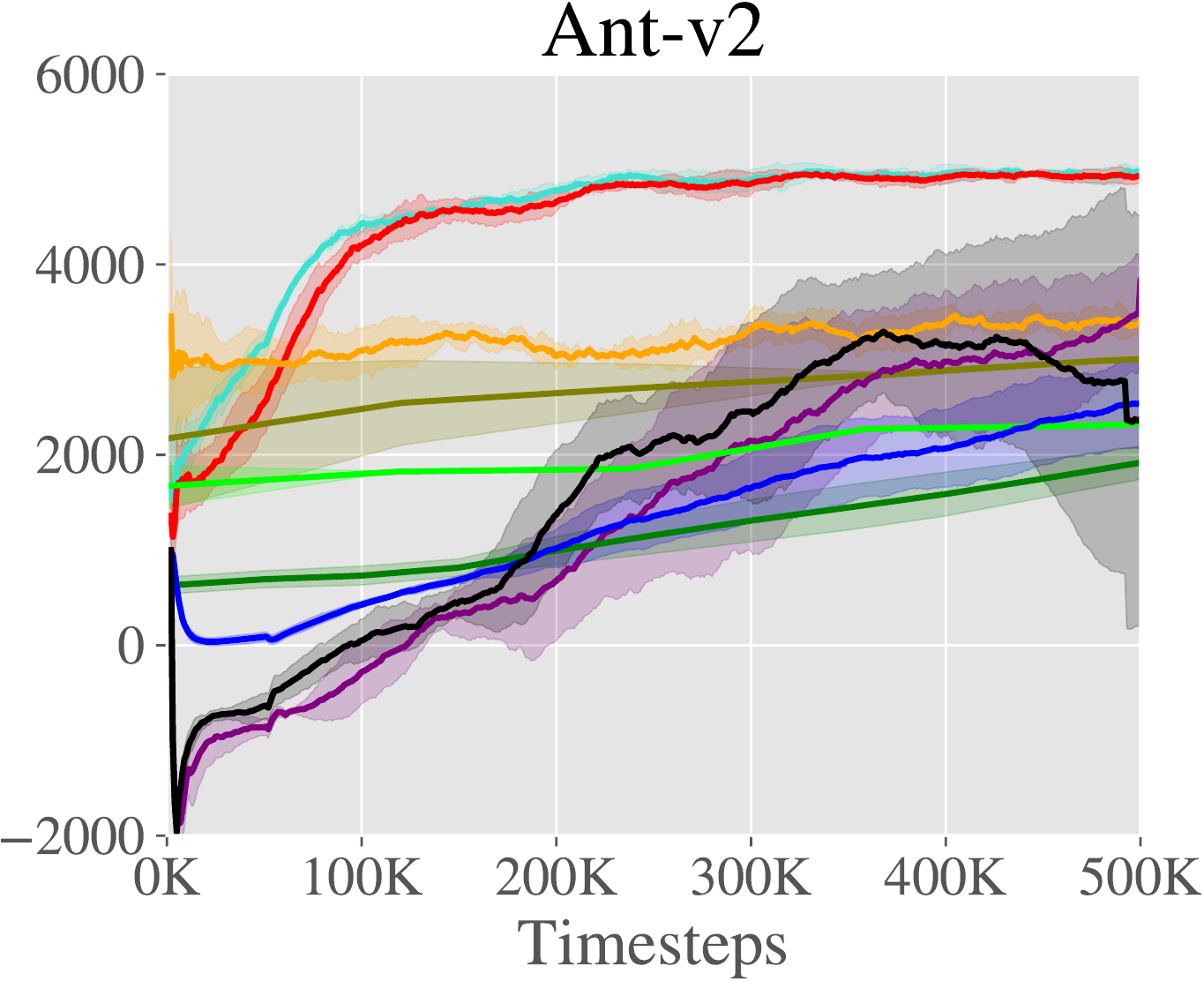}
    \end{subfigure}
    \begin{subfigure}[b]{0.3\textwidth}
        \center
        \includegraphics[width=\textwidth]{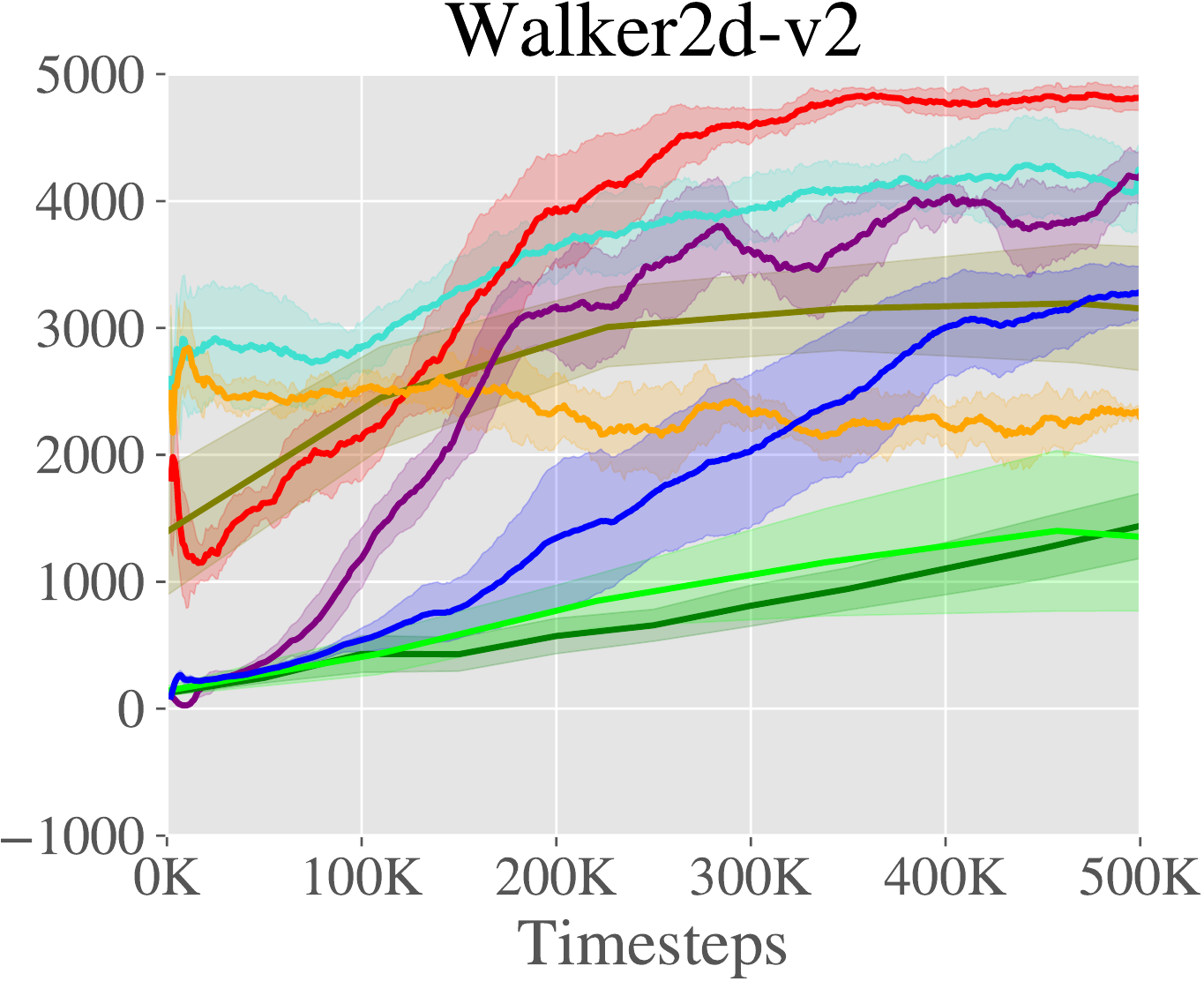}
    \end{subfigure}
    
    \begin{subfigure}[b]{0.99\textwidth}
        \center
        \includegraphics[width=0.6\textwidth]{figures/mujoco/legend_ncol2-crop.pdf}
    \end{subfigure}
    \caption{
    Comparison of our method and prior methods on standard MuJoCo benchmark tasks. These tasks are much easier than the dexterous manipulation tasks, and allow us to better inspect the performance of methods in the setting of offline pretraining followed by online fine-tuning. SAC+BC and BRAC perform on par with our method on the HalfCheetah task, and ABM performs on par with our method on the Ant task, while our method outperforms all others on the Walker2D task. 
    Our method matches or exceeds the best prior method in all cases, whereas no other single prior method attains good performance on all of the tasks. }
    \label{fig:sim-learning-curves}
\end{figure*}

\textbf{Bootstrapping Error Accumulation Reduction (BEAR).} We utilize the implementation of BEAR provided in \href{https://github.com/vitchyr/rlkit}{rlkit}. We provide the demonstration and off-policy data to the method together. Since the original method only involved training offline, we modify the algorithm to include an online training phase. In general we found that the MMD constraint in the method was too conservative. As a result, in order to obtain the results displayed in our paper, we swept the MMD threshold value and chose the one with the best final performance after offline training with offline fine-tuning.



\subsection{Gym Benchmark Results From Prior Data}
\label{sec:gym}

In this section, we provide a comparative evaluation on MuJoCo benchmark tasks for analysis. These tasks are simpler, with dense rewards and relatively lower action and observation dimensionality. Thus, many prior methods can make good progress on these tasks. These experiments allow us to understand more precisely which design decisions are crucial. For each task, we collect 15 demonstration trajectories using a pre-trained expert on each task, and 100 trajectories of off-policy data by rolling out a behavioral cloned policy trained on the demonstrations. The same data is made available to all methods. The results are presented in Figure~\ref{fig:sim-learning-curves}. AWAC is consistently the best or on par with the best-performing method. No other single method consistently attains the best results -- on HalfCheetah, SAC + BC and BRAC are competitive, while on Ant-v2 ABM is competitive with AWAC.
We summarize the results according to the challenges in Section~\ref{sec:challenges}.

\textbf{Data efficiency.} The three methods that do not estimate $Q^\pi$ are DAPG~\citep{rajeswaran2018dextrous}, AWR~\citep{peng2019awr}, and MARWIL~\citep{wang2018marwil}. Across all three tasks, we see that these methods are somewhat worse offline than the best performing offline methods, and exhibit steady but very slow improvement during fine-tuning. In robotics, data efficiency is vital, so these algorithms are not good candidates for practical real-world applications.

\textbf{Bootstrap error in offline learning.} For SAC~\citep{haarnoja2018sac}, across all three tasks, we see that the offline performance at epoch 0 is generally poor. Due to the data in the replay buffer, SAC with prior data does learn faster than from scratch, but AWAC is faster to solve the tasks in general. SAC with additional data in the replay buffer is similar to the approach proposed by~\citet{vecerik17ddpgfd}. SAC+BC reproduces~\citet{nair2018demonstrations} but uses SAC instead of DDPG~\citep{lillicrap2015continuous} as the underlying RL algorithm. We find that these algorithms exhibit a characteristic dip at the start of learning. Although this dip is only present in the early part of the learning curve, a poor initial policy and lack of steady policy improvement can be a safety concern and a significant hindrance in real-world applications. Moreover, recall that in the more difficult dextrous manipulation tasks, these algorithms do not show any significant learning.

\textbf{Conservative online learning.} Finally, we consider  conservative offline algorithms: ABM~\citep{siegel2020abm}, BEAR~\citep{kumar19bear}, and BRAC~\citep{wu2019brac}. We found that BRAC performs similarly to SAC for working hyperparameters. BEAR trains well offline -- on Ant and Walker2d, BEAR significantly outperforms prior methods before online experience. However, online improvement is slow for BEAR and the final performance across all three tasks is much lower than AWAC. The closest in performance to our method is ABM, which is comparable on Ant-v2, but much slower on other domains. 

\pagebreak

\subsection{Extra Baseline Comparisons (CQL, AlgaeDICE)}

\begin{figure}[H]
    \centering
    \begin{subfigure}[b]{0.49\textwidth}
        \center
        \includegraphics[width=\textwidth]{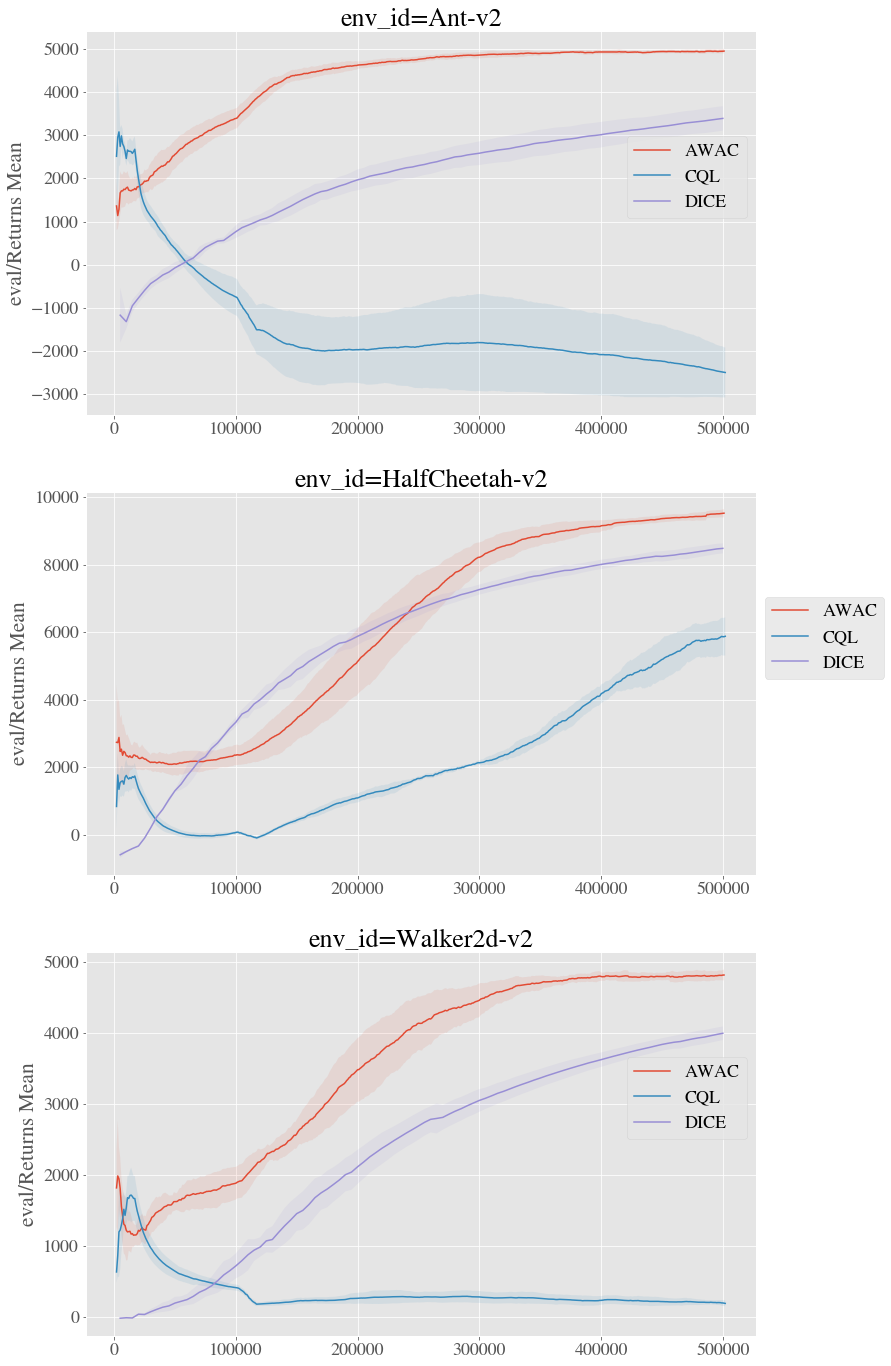}
    \end{subfigure}
    \begin{subfigure}[b]{0.49\textwidth}
        \center
        \includegraphics[width=\textwidth]{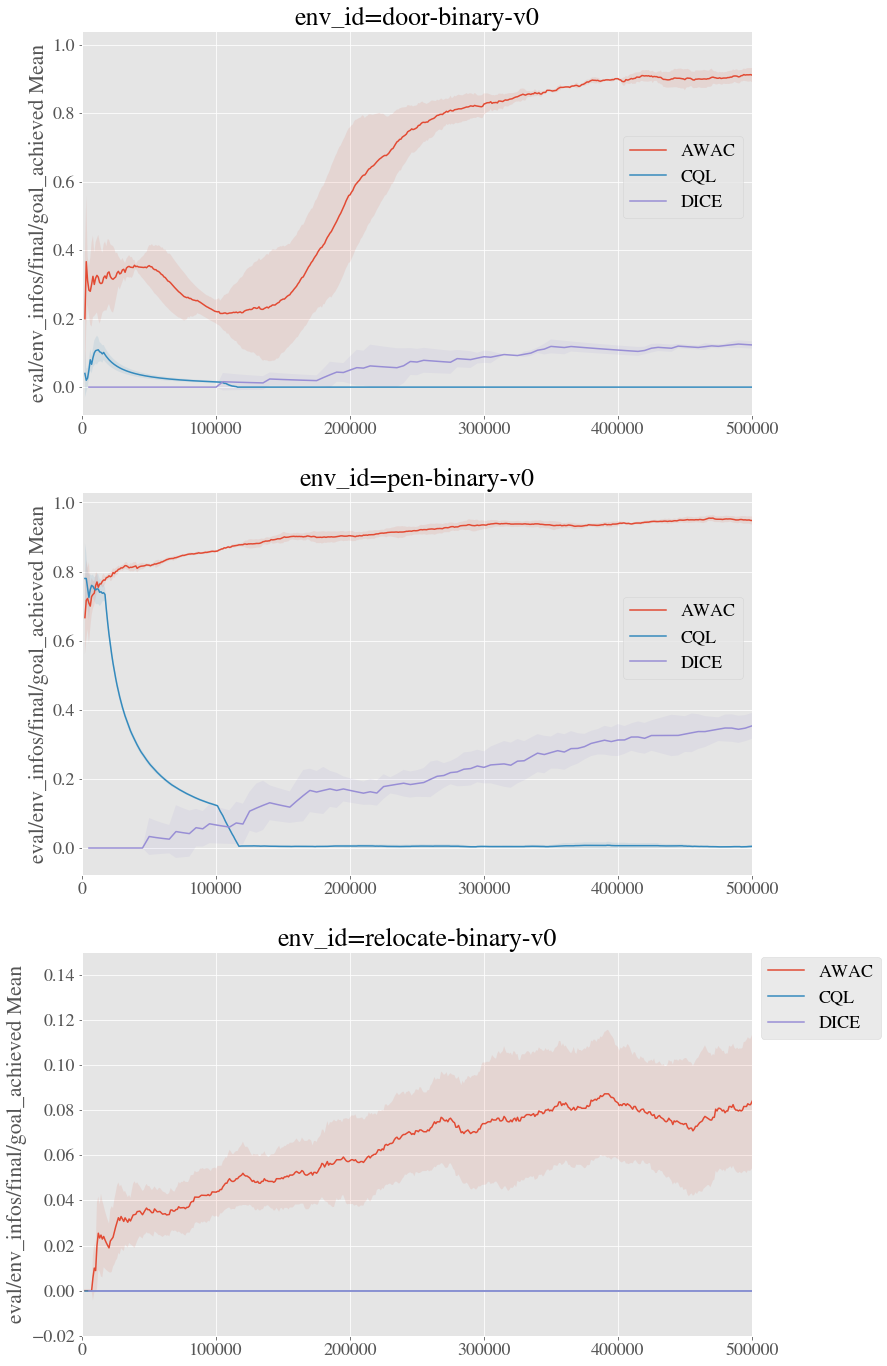}
    \end{subfigure}
    \caption{Comparison of our method (AWAC) with CQL and AlgaeDICE. CQL and AWAC perform similarly offline, but CQL does not improve when fine-tuning online. AlgaeDICE does not perform well for offline pretraining. }
    \label{fig:cql-dice}
\end{figure}

In this section, we add comparisons to constrained Q-learning (CQL)~\citep{kumar2020cql} and AlgaeDICE~\citep{nachum2019dualdice}. For CQL, we use the authors' implementation, modified for additionally online-finetuning instead of only offline training. For AlgaeDICE, we use the publicly available implementation, modified to load prior data and perform 25K pretraining steps before online RL. The results are presented in Figure~\ref{fig:cql-dice}.

\subsection{Online Fine-Tuning From D4RL}

In this experiment, we evaluate the performance of varied data quality (random, medium, medium-expert, and expert) datasets included in D4RL~\citep{fu2020d4rl}, a dataset intended for offline RL. The results are obtained by first by training offline and then fine-tuning online on each setting for 500,000 additional steps. The performance of BEAR~\citep{kumar19bear} is attached as reference. We attempted to fine-tune BEAR online using the same protocol as AWAC but the performance did not improve and often decreased; thus we report the offline performance. All performances are scaled to 0 to 100, where 0 is the average returns of a random policy and 100 is the average returns of an expert policy (obtained by training online with SAC), as is standard for D4RL.

The results are presented in Figure~\ref{fig:d4rl_comparison}. First, we observe that AWAC (offline) is competitive with BEAR, a commonly used offline RL algorithm. Then, AWAC is able to make progress in solving the tasks with online fine-tuning, even when initialized from random data or “medium” quality data, as shown by the performance of AWAC (online). In almost all settings, AWAC (online) is the best performing or tied with BEAR. In four of the six lower quality (random or medium) data settings, AWAC (online) is significantly better than BEAR; it is reasonable that AWAC excels in the lower-quality data regime because there is more room for online improvement, while both offline RL methods often start at high performance when initialized from higher-quality data.

\begin{figure}[H]
\begin{tabular}{lllll}
 &    & \begin{tabular}[c]{@{}l@{}}AWAC\\ (offline)\end{tabular} & \begin{tabular}[c]{@{}l@{}}AWAC\\ (online)\end{tabular} & BEAR \\
HalfCheetah & random & 2.2 & \textbf{52.9}  & 25.5 \\
 & medium & 37.4 & 41.1 & 38.6 \\
 & medium-expert & 36.8 & 41.0 & \textbf{51.7} \\
 & expert & 78.5 & 105.6 & 108.2         \\
Hopper      & random & 9.6 & \textbf{62.8} & 9.5 \\
 & medium & 72.0 & \textbf{91.0} & 47.6 \\
 & medium-expert & 80.9 & \textbf{111.9} & 4.0 \\
 & expert  & 85.2 & 111.8 & 110.3 \\
Walker2D  & random & 5.1 & 11.7 & 6.7 \\
 & medium & 30.1 & \textbf{79.1} & 33.2 \\
 & medium-expert & 42.7 & \textbf{78.3} & 10.8 \\
 & expert & 57.0 & 103.0 & 106.1        
\end{tabular}
\caption{Comparison of our method (AWAC) fine-tuning on varying data quality datasets in  D4RL~\citep{fu2020d4rl}. AWAC is able to improve its offline performance by further fine-tuning online. }
\label{fig:d4rl_comparison}
\end{figure}

\begin{figure*}[t]
    \centering
    \begin{subfigure}[b]{0.32\textwidth}
        \includegraphics[height=0.6\textwidth]{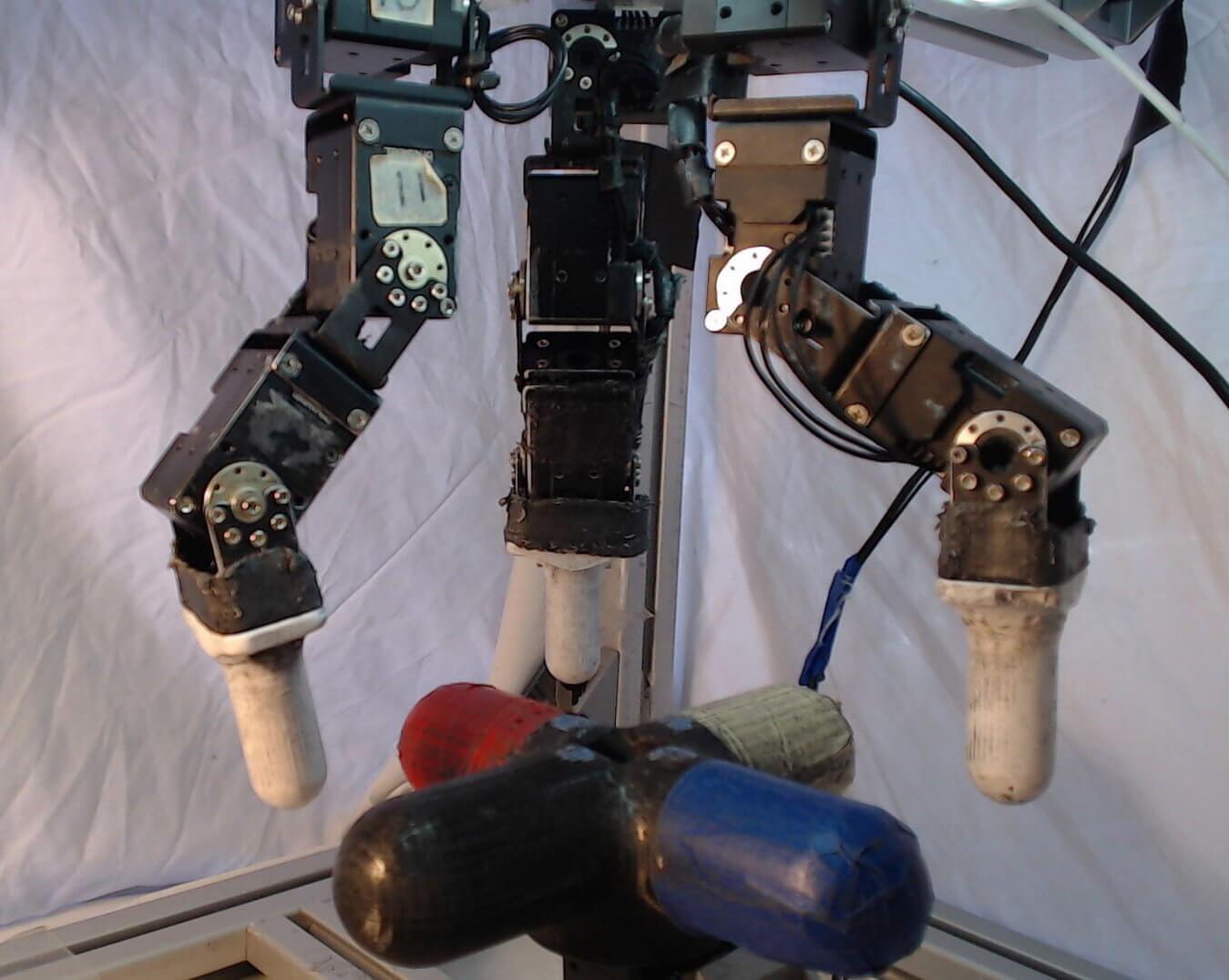}
    \end{subfigure}
    \begin{subfigure}[b]{0.32\textwidth}
        \center
        \includegraphics[height=0.6\textwidth]{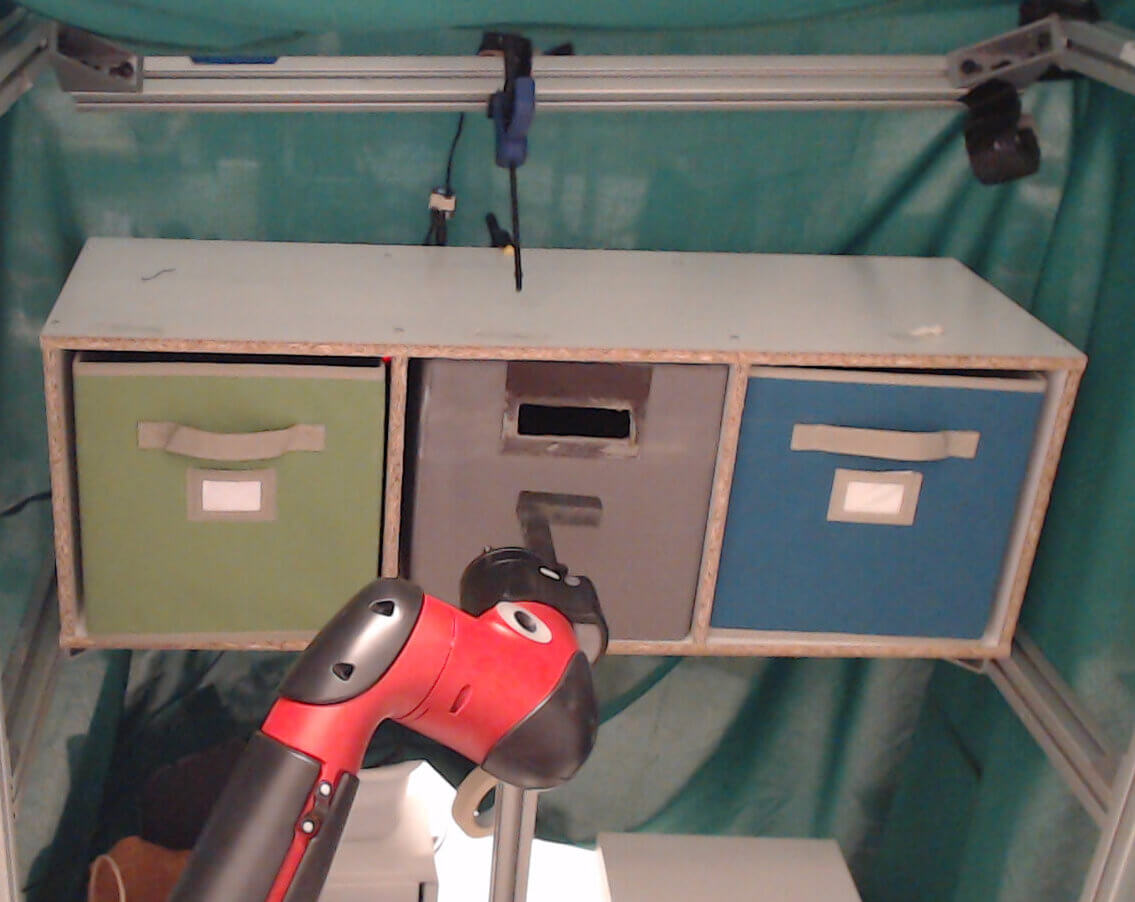}
    \end{subfigure}
    \begin{subfigure}[b]{0.32\textwidth}
        \center
        \includegraphics[height=0.6\textwidth]{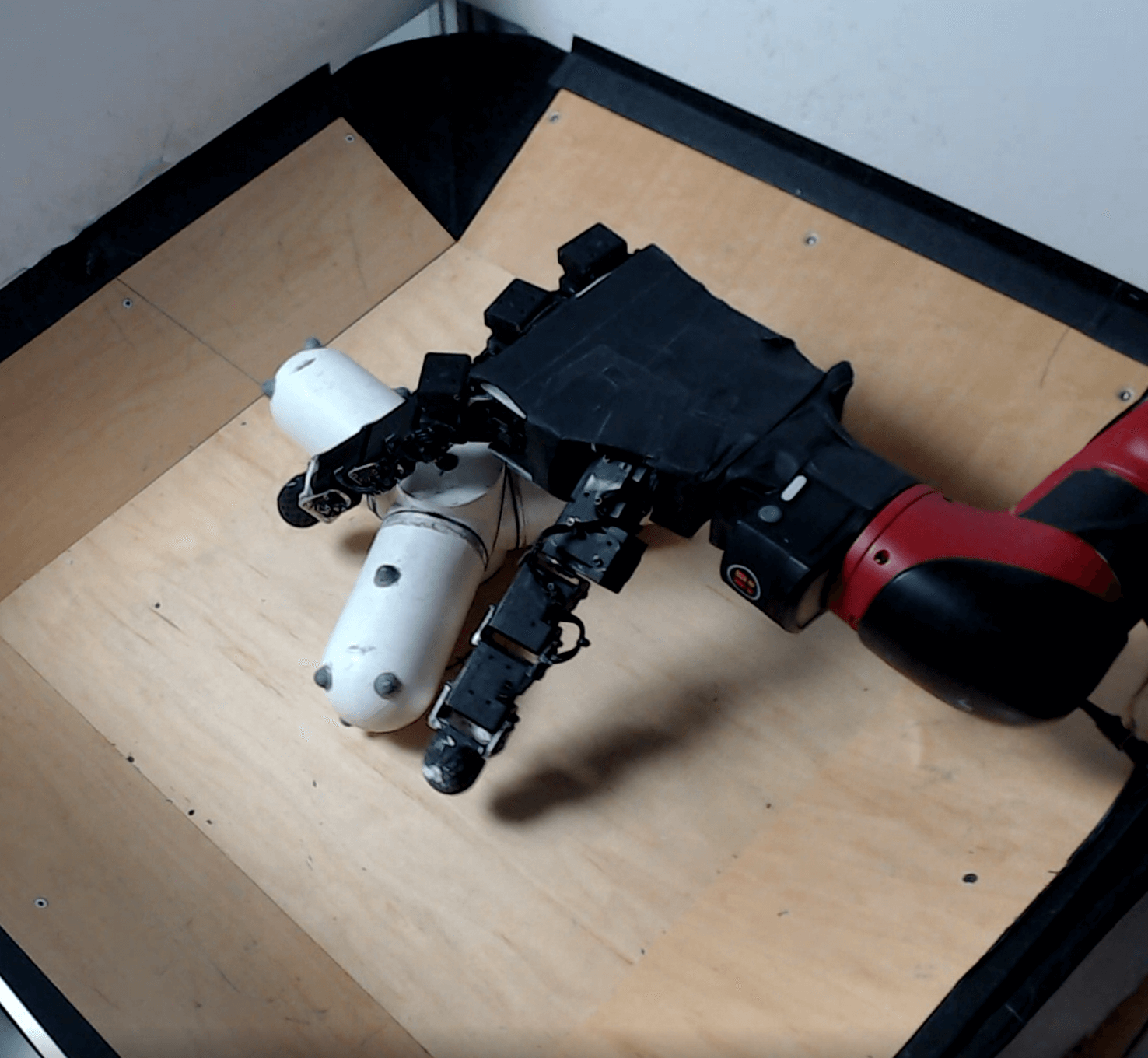}
    \end{subfigure}
    \caption{
    Full views of the robot hardware setups. Videos are available at \projectpage
    }
    \label{fig:robot_setups}
\end{figure*}

\pagebreak

\subsection{Hardware Experimental Setup}
\label{sec:hardwaresetup}

Here, we provide further details of the hardware experimental setups, which are pictured in Fig~\ref{fig:robot_setups}.

\noindent \textbf{Dexterous Manipulation with a 3 Fingered Claw.}
\begin{itemize}
    \item State space: 22 dimensions, consisting of joint angles of the robot and rotational position of the object. 
    \item Action space: 9 dimensions, consisting of desired joint angles of the robot.
    \item Reward: $-1$ if the valve is rotated within 0.25 radians of the target, and $0$ otherwise.
    \item Prior data: 10 demonstrations collected by kinesthetic teaching and 200 trajectories of behavior cloned data.
\end{itemize}

\noindent \textbf{Drawer Opening with a Sawyer Arm.}
\begin{itemize}
    \item State space: 4 dimensions, end effector position of the robot and rotational position of the motor attached to the drawer. 
    \item Action space: 3 dimensions, for velocity control of end-effector position.
    \item Reward: $-1$ if the motor is rotated more than 15 radians of the reset position, and $0$ otherwise.
    \item Prior data: 10 demonstrations collected using a 3DConnexion Spacemouse device and 500 trajectories of behavior cloning data.
\end{itemize}

\noindent \textbf{Dexterous Manipulation with a Robotic Hand.}
\begin{itemize}
    \item State space: 25 dimensions, consisting of joint angles of the hand, end effector positions of the arm, object position and target position. 
    \item Action space: 19 dimensions, consisting of desired 16 joint angles of the hand and 3 dimensions for end-effector control of the arm.
    \item Reward: let $o$ be the position of the object, $h$ be the position of the hand, and $g$ be the target location of the object. Then $r = -||o - h|| - 3||o - g||$.
    \item Prior data: 19 demonstrations obtained via kinesthetic teaching and 50 trajectories of behavior cloned data.
\end{itemize}

\end{document}